\documentclass[11pt]{article}
\usepackage[final]{acl}

\usepackage{times}
\usepackage{latexsym}
\usepackage[T1]{fontenc}
\usepackage[utf8]{inputenc}
\usepackage{microtype}
\usepackage{inconsolata}
\usepackage{graphicx}
\usepackage{tikz}
\usepackage{pgfplots}
\pgfplotsset{compat=1.18} 
\usetikzlibrary{pgfplots.groupplots}
\usepackage{relsize}

\usepackage{amsmath}
\usepackage{booktabs}
\usepackage{multirow}

\newcommand{\Sref}[1]{\S\ref{#1}}

\newcommand{\Fref}[1]{Figure~\ref{#1}}

\newcommand{\Tref}[1]{Table~\ref{#1}}
\newcommand\Modern{\textsc{modern}}
\newcommand\Historical{\textsc{historical}}
\newcommand\pmodern{$\mathcal{D}^{Pub}_{\textsc{modern}}$}
\newcommand\phistorical{$\mathcal{D}^{Pub}_{\textsc{historical}}$}
\newcommand\tmodern{$\mathcal{D}^{Twt}_{\textsc{modern}}$}
\newcommand\thistorical{$\mathcal{D}^{Twt}_{\textsc{historical}}$}

\newcommand\WordForm{\texttt}

\title{From \emph{sunblock} to \emph{softblock}: Analyzing the correlates of neology in published writing and on social media}

\author{
 \textbf{Maria Ryskina\textsuperscript{1$\ddagger$}},
 \textbf{Matthew R. Gormley\textsuperscript{2}},
 \textbf{Kyle Mahowald\textsuperscript{3}},
 \textbf{David R. Mortensen\textsuperscript{2}},
\\
 \textbf{Taylor Berg-Kirkpatrick\textsuperscript{4}},
 \textbf{Vivek Kulkarni}
 \\
 \\
 \textsuperscript{1}Vector Institute for Artificial Intelligence,
 \textsuperscript{2}Carnegie Mellon University,\\
 \textsuperscript{3}The University of Texas at Austin,
 \textsuperscript{4}University of California, San Diego
\\ 
\small{
\textsuperscript{$\ddagger$}Work done partly at Carnegie Mellon University
} \\
 \small{
    \textbf{Correspondence:} \href{mailto:mryskina@alumni.cmu.edu}{mryskina@alumni.cmu.edu}
  }
}

\begin{document}
\maketitle
\begin{abstract}

Living languages are shaped by a host of conflicting internal and external evolutionary pressures. While some of these pressures are universal across languages and cultures, others differ depending on the social and conversational context: language use in newspapers is subject to very different constraints than language use on social media. Prior distributional semantic work on English word emergence \emph{(neology)} identified two factors correlated with creation of new words by analyzing a corpus consisting primarily of historical published texts \cite{ryskina-etal-2020-new}. Extending this methodology to contextual embeddings in addition to static ones and applying it to a new corpus of Twitter posts, we show that the same findings hold for both domains, though the topic popularity growth factor may contribute less to neology on Twitter than in published writing.
We hypothesize that this difference can be explained by the two domains favouring different neologism formation mechanisms.\footnote{Code, word lists, and tweet IDs can be found at \\
\url{https://github.com/ryskina/twitter-neology/}}
\end{abstract}

\section{Introduction}

Hypotheses about the mechanisms of language evolution have traditionally been tested by comparing linguistic data over long periods of time \cite{aitchison2001language}. Until the end of the last century, the majority of the textual sources preserved across many decades have been books, periodicals, or other forms of published writing. Though analyses of such texts can yield important discoveries,
they do not represent the entirety of their time's everyday language use. 
Institutions like publishers and news agencies tend to favour standardization  \cite{milroy2012} and adopt linguistic innovation less readily than individuals: by the time a word appears in print, it is likely already widely used.\footnote{\citet[p. 255]{rogers1971communication} conclude that adoption of innovation in general is more likely to spread through interpersonal channels than through mass media.} Historical print media data is also unlikely to reflect the language of underrepresented and marginalized communities---a fertile ground for linguistic creativity \cite{montgomery2008introduction,lefkowitz2017anti}.

The Internet has opened up new avenues for observing and studying language change \cite{mcculloch2020because}. Chats, blogs, and social media contain unprecedented amounts of linguistic data and represent a greater diversity of voices and styles. This richness makes online communication especially well-suited for population-level analysis of the evolutionary forces \cite{bowern-2019-semantic-2} that drive innovation, such as social prestige, cognitive economy, or communicative need---the language users' collective desire to express certain meanings.

We have previously tested two hypotheses formalizing the role of communicative need in word emergence (\emph{neology}; \citealp{ryskina-etal-2020-new}). 
Analyzing a historical corpus primarily comprised of books
and other publications, we found that new words are created both to give names to new concepts in domains of growing cultural importance and to fill in gaps in the word distribution over the space of possible meanings, with the former factor being more significant. In this paper, we ask: will an analysis of the same factors in social media neology yield similar findings? And if there are differences, are they simply 
due to
the limitations of historical published corpora (e.g., sparsity), or are the underlying pressures of word emergence different between the two domains? 

We collect a corpus of 260M English-language tweets (\Sref{sec:twitter-data}) and extend our original methodology \cite{ryskina-etal-2020-new} to study the correlates of neology both on social media and in published writing. Experimenting with different operationalizations (e.g., using both static and contextual word embeddings; \Sref{sec:embeddings}) of our two hypotheses (\Sref{sec:hypotheses}), we robustly reproduce our earlier findings for a corpus of published writing (support for both hypotheses; \Sref{sec:results-cocacoha}). We then perform the same analysis for Twitter and observe similar trends (\Sref{sec:results-twitter}), though frequency growth of the neighbouring words may play a less important role there.

\section{Related work}

\paragraph{NLP studies of neology and word decline}

Several computational studies have focused on the mechanisms and driving factors of word emergence (and loss) in languages, building on cognitive and linguistic models \cite{xu2019predictability,xu2023predicting,karjus2020quantifying,karjus2021conceptual}. One line of work uses this knowledge to trace the 
etymological origins of neologisms \cite{kulkarni-wang-2018-simple,wu-yarowsky-2020-computational}, especially lexical blends \cite{cook-stevenson-2010-automatically-identifying, pinter-etal-2020-will}. Other studies model the process as a whole, predicting what makes words or senses likely to emerge \cite{hofmann-etal-2020-predicting-2, sun-xu-2022-tracing}, persist \cite{stewart-eisenstein-2018-making}, or decline \cite{francis-etal-2021-quantifying,kali2024cognitive}. Most relevant to the current study is our prior work \cite{ryskina-etal-2020-new} which tests two hypotheses about intra- and extra-linguistic pressures that facilitate word emergence (\Sref{sec:hypotheses}); this study builds primarily on the experimental methodology introduced there.

Processing of novel words remains a significant challenge in NLP applications. Neologisms, along with synthetic nonce words, are used to test model generalization in domains like translation \cite{zheng-etal-2024-neo, lerner-yvon-2025-towards}, morphology \cite{weissweiler-etal-2023-counting-2, mortensen-etal-2024-verbing}, and definition generation \cite{malkin-etal-2021-gpt}.

\paragraph{Diachronic analysis of published writing}

Large time-stamped corpora of texts lend themselves to comparative analyses of historical language change. Such corpora typically span decades or centuries and include literature \cite{michel2011quantitative,bouma-etal-2020-edges}, news articles \cite{popescu-strapparava-2015-semeval,basile-etal-2020-diachronic}, or both \cite{onelli-etal-2006-diacoris,davies2009coca,davies2012coha}.

In the computational linguistics community, such corpora have most often been used for tracking the evolution of word meanings \cite{kutuzov-etal-2018-diachronic, tahmasebi2021survey}, typically quantified by measuring changes in word embeddings across time. Earlier approaches included learning static embeddings from different time periods \cite{kulkarni2015statistically,hamilton-etal-2016-diachronic} or building temporally-aware dynamic embeddings \cite{del2016tracing,10.1145/3178876.3185999}, and the rise of pretrained language models made contextual embeddings more popular (e.g., \citealp{giulianelli-etal-2020-analysing, martinc-etal-2020-leveraging, montariol-etal-2021-scalable,periti-tahmasebi-2024-systematic}; see \citealp{10.1145/3672393} for a survey).

Diachronic book corpora are also used for evaluating text-dating models \cite{kulkarni-etal-2018-simple,ren-etal-2023-time}, analyzing change in syntactic phenomena \cite{krielke-etal-2022-tracing,miletic-schulte-im-walde-2025-modeling},
and applications in digital humanities \cite{ruiz-etal-2017-enjambment,klaussner-vogel-2018-diachronic,haider-eger-2019-semantic-2}.

\paragraph{Diachronic analysis of social media}

Social media is an exceptionally rich domain for studying language change. Users adopt novel words and spellings to express their creativity, signal community membership, or avoid moderation in scenarios ranging from protecting marginalized users from censorship \cite{ungless2025lesbean} to disguising dog whistles \cite{kruk-etal-2024-silent,sasse-etal-2025-making}.

Most existing quantitative work on language change in social media focuses on tracking the life cycle of emergent words \cite{10.1145/2488388.2488416,wurschinger-etal-2016-using,stewart-eisenstein-2018-making,10.3389/frai.2021.648583} and the social, demographic, and geographical factors that affect it \cite{10.1371/journal.pone.0113114,grieve2018mapping,ananthasubramaniam2024networks,10.1145/3696410.3714716}. To our knowledge, ours is the first study of the semantic factors driving word emergence on social media.

\section{Question and hypotheses \label{sec:hypotheses}}

We define `neologism' broadly as a novel form--meaning pair that at some point enters more general use (as reflected by a substantial increase in the form's usage frequency; \citealp{ryskina-etal-2020-new}). This includes new coinages \emph{(yeet)}, combinations of known elements \emph{(tiktoker, cryptocurrency)}, or existing words paired with new senses \emph{(transformer)}. In particular, we are interested in \emph{what kinds of meanings} such neologisms are likely to express.

We reuse the supply-driven and demand-driven neology hypotheses introduced in our prior work \cite{ryskina-etal-2020-new}.
These competing hypotheses rely on the concept of a semantic space---a manifold of meanings where certain points correspond to words---and posit where in such a space neologisms are likely to emerge. We operationalize the hypotheses under the distributional semantics paradigm \cite{lenci2023distributional}, using word embedding spaces informed by the co-occurrence statistics within our corpora as a proxy for the underlying semantic space.

\paragraph{Supply hypothesis} This hypothesis suggests that neologisms are more likely to emerge in \emph{sparser areas of the semantic space.} It is derived from \citeposs{breal1904essai} law of differentiation, which states that the semantic space tends towards uniformity; if two existing words are too close in meaning, they will either diverge or one of them will fall out of use. By extension, we posit that the existence of gaps in the semantic space could create pressure on the language to repair uniformity by populating them with neologisms.

\paragraph{Demand hypothesis} The second hypothesis states that neologisms are more likely to emerge in \emph{semantic neighbourhoods of growing popularity.} Growing popularity of a certain semantic cluster---i.e., growing frequency of use for the words that make it up---can be viewed as a reflection of the increasing importance of the corresponding domain of discourse. Rapidly developing domains such as technology might produce novel concepts faster, and the need for words to refer to these new  concepts could also be a driving factor of neology.

\section{Data \label{sec:data}}

\begin{table*}[t]
    \centering
    \begin{tabular}{lcccccc}
    \toprule
    & Texts & Sentences & Tokens & Types & Avg. text length & Avg. sentence length \\ 
    \midrule
    \phistorical{} & \hphantom{1}93K & \hphantom{1}21M & \hphantom{1,}385M & 1.8M & 4,140 tokens & 18 tokens \\
    \pmodern{} & 151K & \hphantom{1}24M & \hphantom{1,}406M & 2.5M & 2,704 tokens & 17 tokens \\
    \midrule
    \thistorical{} & \hphantom{1}31M & \hphantom{1}48M & \hphantom{1,}409M & 3.0M & \hphantom{1,1}13 tokens & \hphantom{1}9 tokens\\
    \tmodern{}   & 229M & 331M & 3,206M & 9.6M & \hphantom{1,1}14 tokens & 10 tokens \\
    \bottomrule
    \end{tabular}
    \caption{Corpus statistics for the published writing data (rows 1--2) and the Twitter data (rows 3--4). The data for each domain is split into a \Historical{} and a \Modern{} subset as described in \Sref{sec:data}. For the published writing domain, `text' = book, story, or article; for Twitter, `text' = tweet. All texts are split into sentences using the NLTK toolkit \cite{bird2009natural}. \pmodern{} and \phistorical{} data is already split at the token level \cite{davies2009coca,davies2012coha}, and we use NLTK's Twitter-specific tokenizer for tweets.
    The `Tokens' column shows the total number of tokens in each subcorpus, while the `Types' column shows the number of unique tokens in each subcorpus.
    }
    \label{tab:corpora}
\end{table*}

\subsection{Published writing corpora \label{sec:cocacoha-data}}

As in our \citeyear{ryskina-etal-2020-new} work, we consider two collections of American English texts from non-overlapping time periods: \Historical{} (1800--1989) and \Modern{} (1990--2012). We will denote these two datasets as \phistorical{} and \pmodern{} respectively. Both are drawn from existing multi-genre diachronic corpora: COHA \citep{davies2012coha} for \phistorical{} and  COCA \citep{davies2009coca} for \pmodern{}. In both corpora, all texts are already annotated for genre and time period (decade or year for COHA and COCA respectively), preprocessed, and tokenized; we do not perform any additional preprocessing besides sentence-splitting. In order to be able to estimate the frequency trends in \phistorical{}, we split this dataset into decades using the temporal labels provided with the COHA corpus. We restrict both datasets to published materials only: fiction, non-fiction, academic articles, and popular magazine and newspaper articles. The statistics for both \phistorical{} and \pmodern{} are reported in \Tref{tab:corpora}.

\subsection{Twitter data collection \label{sec:twitter-data}}

We collect a corpus of $\sim$260M tweets using the Twitter API,\footnote{Last accessed in January 2023.} spanning the period from January 
2007
to December 
2021. Aiming for uniform coverage, we randomly sample 100K unique tweets per day; if the total number of tweets available for a given day is <100K  (e.g., in the early days of the platform's existence), we use the entire set. We restricted the tweets to be in English only and excluded retweets or tweets from bots. Since there were much fewer users on the platform in the initial years after its launch in 2007, the tweet distribution in our corpus is skewed: the number of tweets per year grows from $\sim$80K in 2007--2008 to 7M in 2009 to 18--24M in 2010--2021. We discard tweets that contain only URLs or other special tokens (0.3\% of tweets total; details in Appendix \ref{sec:appendix-tokenization}). 

We split the collected data into the \Historical{} (representing the ``baseline'' distribution of word use) and \Modern{} (in which the neologisms appear) subsets, denoted \thistorical{} and \tmodern{}. All tweets are labelled by year; we allocate tweets from 2007--2010 to the \Historical{} subset and tweets from 2011--2021 to the \Modern{} subset. The statistics for both \thistorical{} and \tmodern{} are reported in \Tref{tab:corpora}.

\section{Methodology \label{sec:methods}}

We build upon the methodology proposed in our previous study \cite{ryskina-etal-2020-new}, with several modifications detailed in this section and in Appendix \ref{sec:appendix-diff}. First, neologisms are identified automatically and filtered (\Sref{sec:selection}). Second, each of them is paired with a non-neologism control word that is similar to the neologism in several aspects (\Sref{sec:control}). Finally, we compare the distribution of the statistics of interest across the neighbours of the neologisms and the control words (\Sref{sec:setup}).

\subsection{Neologism selection \label{sec:selection}}

\paragraph{Published writing \label{sec:selection-cocacoha}} We reuse the list of neologisms extracted from the COCA-based \pmodern{} data used in our \citeyear{ryskina-etal-2020-new} study. It includes 1000 nouns that appear substantially more frequently in the \pmodern{} split than in the \phistorical{} one.\footnote{Note that the \pmodern{} corpus from the \citeyear{ryskina-etal-2020-new} study is a superset of the one in this work: it included speech transcripts, excluded from this study due to focus on published writing.}

\paragraph{Twitter \label{sec:selection-twitter}} Since social media data exhibits much greater lexical diversity than published works, simply comparing frequencies in \thistorical{} and \tmodern{} results in a noisy set of candidates.
Instead, we follow the method of \citet{kulkarni-etal-2018-simple}, which for a given word estimates the year when it came into popular usage. Given a diachronic corpus spanning timesteps $\{1, \ldots, T\}$ (here corresponding to years $y \in \{2007, \ldots, 2021\}$, i.e., $t = y - 2006$) and a word $w$, they find the timestep $t$ when the cumulative usage of $w$ first exceeds a specific percentage of its total cumulative usage through the entire corpus:
\begin{equation*}
    t^* = \arg\min_{t} \left[ \sum_{i=1}^{t}c_i(w) > \alpha \cdot \sum_{j=1}^{T}c_j(w) \right],
\end{equation*}
where $c_t(w)$ represents how many times the word $w$ was used in the timestep $t$. Words which came into popular usage during the \Modern{} period $(y^* = t^* + 2006 \geq 2011)$ are selected as potential neologisms. Empirically we set $\alpha=1/300$.

We further refine the resulting list using part-of-speech tagging (e.g., to remove proper nouns) and frequency-based heuristics (to remove rare variants or filter out auto-generated tweets); see Appendix \ref{sec:appendix-pos} for details. Unlike our \citeyear{ryskina-etal-2020-new} study, here we do not restrict the neologisms to nouns only. After this step, 938 potential neologisms are left.

\paragraph{Manual filtering \label{sec:selection-filtering}} For a stricter analysis, we use dictionaries and other resources to filter out words that would not be considered neologisms per our definition (\Sref{sec:hypotheses}).  For the published writing domain, we remove candidate neologisms that have been in use prior to 1900 and have not gained new senses since then. For Twitter, we manually classify all candidate neologisms to remove all proper names, foreign words, typos and errors, and words that have been in use before 2000 and have not gained new senses since. The details of the filtering procedure are described in Appendix \ref{sec:appendix-filtering}. The results reported in the following sections are based on this filtered set of neologisms; the same for the non-filtered word list can be found in Appendix \ref{sec:appendix-full}.

\subsection{Embeddings \label{sec:embeddings}}

One common approach in modelling semantic shift is learning separate sets of static word embeddings from subcorpora that represent different time periods and then projecting them into a common space for comparison \cite{tahmasebi2021survey}. Recently, contextual embeddings extracted from BERT \cite{devlin-etal-2019-bert} or related models have replaced static embeddings as the tool of choice \cite{periti-tahmasebi-2024-systematic}.
We experiment with both types to test our word emergence hypotheses. 

\paragraph{Static embeddings \label{sec:embeddings-word2vec}} We train separate Word2Vec SkipGram embeddings (\citealp{mikolov2013distributed}; window size $=5$, dimension $=300$) on the \Historical{} and the \Modern{} subcorpora. In prior work on language change, static embedding spaces from distinct time periods are often projected into the same axes via a linear transformation, which seeks to align a set of anchor points between the two spaces \cite{kulkarni2015statistically,zhang-etal-2015-omnia,7511732,hamilton-etal-2016-diachronic}. Specifically, we use an orthogonal Procrustes transformation (following the setup of \citealp{hamilton-etal-2016-diachronic}) with all words that exist in the vocabularies of both models used as anchors. The alignment step is necessary for finding 
the neighbourhoods in the \Historical{} space where neologisms eventually appear: as most neologisms are not in the vocabulary of the \Historical{} Word2Vec model, we approximate their positions in the \Historical{} space by projecting their \Modern{} vectors into the \Historical{} axes. 

We set the minimum word frequency threshold for both the \Historical{} and the \Modern{}  vocabularies to keep them under 100K words (resulting in $\sim$98K vocabulary words for \phistorical{}, $\sim$99K words for \pmodern{}, and $\sim$100K words for \thistorical{} and \tmodern{}). These vocabularies, denoted $V_H$ and $V_M$, are used for the rest of the experiments with both static and contextual embeddings.

\paragraph{Contextual embeddings \label{sec:embeddings-roberta}} These embeddings are richer and better for capturing sense variation. However, having many embeddings for the same word makes it difficult to define the ``neighbour'' relationship between words, which is central for our hypotheses. To avoid this, we average the embeddings to reduce them to their static versions \cite{bommasani-etal-2020-interpreting-2}.

We extract the 768-dimensional contextual embeddings from the RoBERTa model, pretrained on a large corpus of English-language books, news articles, Wikipedia articles, and other online content \cite{liu2019roberta}. During pretraining, 15\% of the words in each sentence were randomly masked, and the model was trained to predict the masked words based on their bidirectional context. We download the pretrained 12-layer RoBERTa-Base model from the Hugging Face Hub.\footnote{\url{https://huggingface.co/FacebookAI/roberta-base}}

We follow the procedure of \citet{timkey-van-schijndel-2021-bark} for obtaining the embeddings. For each of the $\sim$99--100K \Historical{} vocabulary words we sample 250 context sentences from $\mathcal{D}^{\{Pub, Twt\}}_{\textsc{historical}}$.\footnote{Some vocabulary words have fewer/no contexts because of tokenization mismatches between RoBERTa and the tokenizer used to create the vocabulary (Appendix \ref{sec:appendix-tokenization}).}
Similarly, for each neologism word we sample 500 context sentences from $\mathcal{D}^{\{Pub, Twt\}}_{\textsc{modern}}$. Each word's embedding in each sampled context (mean-pooled over subword tokens) is extracted from the last layer of RoBERTa-Base. We z-score all obtained vectors to get rid of the ``rogue dimensions'' that affect word vector similarity \cite{timkey-van-schijndel-2021-bark}. Finally, we average each word's embeddings across its sampled contexts to obtain one static, decontextualized vector per word.

\subsection{Control set selection \label{sec:control}}

To see how the neighbourhoods of neologisms differ from the neighbourhoods of other similar non-neologism words, we pair each neologism with a control word \cite{dubossarsky-etal-2017-outta}. As in prior work, we control for word frequency and length \cite{ryskina-etal-2020-new,francis-etal-2021-quantifying}, and additionally constrain the control to be semantically similar to the neologism. Formally, for each neologism $w_n$ we select a counterpart $w_c$ satisfying the following constraints:
\begin{itemize}
    \item Frequency ranks of the two words in the corresponding corpora are in the same percentile:
    $\left|\frac{z_M(w_n)}{|V_M|} - \frac{z_H(w_c)}{|V_H|} \right| \leq 0.01 $. Here $1 \leq z_H(\cdot) \leq |V_H|$ and $1 \leq z_M(\cdot) \leq |V_M|$ are ranks of the words in the \Historical{} and \Modern{} vocabularies, sorted by frequency;
    \item The length of the two words is identical up to 3 characters;
    \item The cosine similarity between the neologism and its control counterpart in the \Historical{} embedding space is above a certain threshold: $\text{cosine} \left( v_{w_n}, v_{w_c} \right) \geq 0.4$. Since we want to keep the neologism--control pairs consistent in all experiments, we only use Word2Vec embeddings for this pairing step. Here $v_w$ denotes a projected \Modern{} static embedding if $w$ is a neologism or a \Historical{} static embedding if $w$ is a control word.
\end{itemize}

We use a maximum bipartite matching algorithm \cite{hopcroft1973n} to pair neologisms and controls, finding matches for 231 of the 459 Twitter neologisms and 557 of the 746 published writing neologisms. After this step, neologisms and controls are only compared as sets; we do not perform any statistical comparison within individual pairs.

\begin{table}[th]
    \centering
    \begin{tabular}{p{2.5cm}cc}
        \toprule
        Domain & Neologism & Control \\
        \midrule
        \multirow{4}{2.5cm}{Published writing (1810--2012)} & 
        \WordForm{e-mail} & \WordForm{message} \\ 
        & \WordForm{sunblock} & \WordForm{nicotine} \\ 
        & \WordForm{sitcom} & \WordForm{cinema}\\
        & \WordForm{dysfunction} & \WordForm{functional}\\
        \midrule
        \multirow{4}{2.5cm}{Tweets (2007--2021)} & \WordForm{cringiest} & \WordForm{silliest} \\
        & \WordForm{softblock} &
        \WordForm{un-followed} \\
        & \WordForm{bruhhhhh} & \WordForm{niceeeee} \\
        & \WordForm{baecation} & \WordForm{staycation} \\
        \bottomrule
    \end{tabular}
    \caption{\label{tab:pair-examples} Example neologism--control word pairs extracted from either corpus.}
\end{table}

\begin{figure*}[tb]
\centering
\begin{tikzpicture}
\pgfplotsset{small, every tick label/.append style={font=\footnotesize}}
\begin{groupplot}[group style = {group size = 3 by 2, horizontal sep = 0.8cm, vertical sep = 0.7cm}, width = 6cm, height = 5cm, ]
    \nextgroupplot[title = {\larger{Neighbourhood density}},
        legend style = { column sep = 10pt, legend columns = -1, legend to name = grouplegend},
        x tick label style={rotate=45,anchor=north east,/pgf/number format/.cd, fixed, fixed zerofill,
        precision=3},
        ybar,
        bar width = 4pt,
        ytick = {0,2,4,6,8,10},
        ymajorgrids=true,
        ymin=0, ymax=10,
        x dir=reverse,
        xtick={0.55,0.525,0.5,0.475,0.45,0.425,0.4,0.375,0.35},
        xticklabel=\empty
        ]
        \addplot+[error bars/.cd,y dir=both,y explicit]
        coordinates {
(0.55, 2.072425257332915) +- (0.0, 0.07867618177956234)
(0.525, 2.756753562994494) +- (0.0, 0.0798141255989022)
(0.5, 3.481988644271132) +- (0.0, 0.07612058075361629)
(0.475, 4.205543723600405) +- (0.0, 0.0689433133486728)
(0.45, 4.879595694137123) +- (0.0, 0.061480003362850116)
(0.425, 5.503056147166612) +- (0.0, 0.05445863323218277)
(0.4, 6.091771271046782) +- (0.0, 0.047036555437317204)
(0.375, 6.638180787975625) +- (0.0, 0.04109814440628057)
(0.35, 7.148854137075472) +- (0.0, 0.035984959927399425)
    }; 
    \addlegendentry{Neighbourhoods of neologisms}%
        \addplot+[error bars/.cd,y dir=both,y explicit] 
        coordinates {
(0.55, 4.363248665331795) +- (0.0, 0.06742406792550916)
(0.525, 4.924514157321661) +- (0.0, 0.062278827396129076)
(0.5, 5.4524720035106755) +- (0.0, 0.05732071940716011)
(0.475, 5.949294101278805) +- (0.0, 0.05163717669095723)
(0.45, 6.419368545941398) +- (0.0, 0.046198414946277466)
(0.425, 6.859529558799917) +- (0.0, 0.04151194579239422)
(0.4, 7.278523588495004) +- (0.0, 0.03710268963682268)
(0.375, 7.678881976280493) +- (0.0, 0.03307204545197982)
(0.35, 8.064816979306777) +- (0.0, 0.02924300292201833)
        }; \addlegendentry{Neighbourhoods of control words}
    \nextgroupplot[title={\larger{Growth monotonicity}},
        x tick label style={rotate=45,anchor=north east,/pgf/number format/.cd, fixed, fixed zerofill,
        precision=3},        
        ybar,
        bar width = 4pt,
        ytick = {0,0.2,0.4,0.6,0.8,1},
        ymajorgrids=true,
        ymin=0, ymax=1,
        x dir=reverse,
        xtick={0.55,0.525,0.5,0.475,0.45,0.425,0.4,0.375,0.35},
        xticklabel=\empty
        ]
        \addplot+[error bars/.cd,y dir=both,y explicit]
        coordinates {
(0.55, 0.7853280237553126) +- (0.0, 0.012872250142418606)
(0.525, 0.7814195180848289) +- (0.0, 0.01306665632173689)
(0.5, 0.7885590451033195) +- (0.0, 0.013201562282321546)
(0.475, 0.7999842981189896) +- (0.0, 0.013396345664097515)
(0.45, 0.8049554651965622) +- (0.0, 0.01364104853044505)
(0.425, 0.826724271139509) +- (0.0, 0.012757937114978761)
(0.4, 0.8201592268769929) +- (0.0, 0.01348744977432379)
(0.375, 0.8219304730301834) +- (0.0, 0.013651336765000964)
(0.35, 0.8262251891212877) +- (0.0, 0.013743131912127775)
        };
        \addplot+[error bars/.cd,y dir=both,y explicit]
        coordinates {
(0.55, 0.6871368521186926) +- (0.0, 0.02121486580509467)
(0.525, 0.6921200010944473) +- (0.0, 0.020733377990943547)
(0.5, 0.6963795578149593) +- (0.0, 0.020268202125799342)
(0.475, 0.7094297723098875) +- (0.0, 0.02013900921647198)
(0.45, 0.7180754187718743) +- (0.0, 0.019665742585232196)
(0.425, 0.7284027472843054) +- (0.0, 0.019445503900397763)
(0.4, 0.7248306106908415) +- (0.0, 0.019978831206548114)
(0.375, 0.7135657815994205) +- (0.0, 0.020480527974013202)
(0.35, 0.7107606909342211) +- (0.0, 0.020486687811931633)
        };
    \nextgroupplot[title={\larger{Growth slope}},
        x tick label style={rotate=45,anchor=north east,/pgf/number format/.cd, fixed, fixed zerofill,
        precision=3},        
        ybar,
        bar width = 4pt,
        ytick = {0,5e-8,1e-7,15e-8,2e-7,25e-8,3e-7},
        ymajorgrids=true,
        ymin=0, ymax=3e-7,
        x dir=reverse,
        xtick={0.55,0.525,0.5,0.475,0.45,0.425,0.4,0.375,0.35},
        xticklabel=\empty
        ]
        \addplot+[error bars/.cd,y dir=both,y explicit]
        coordinates {
(0.55, 2.1456398815789284e-07) +- (0.0, 3.922619279844098e-08)
(0.525, 1.649877356915317e-07) +- (0.0, 2.373681932480172e-08)
(0.5, 1.4538210941200447e-07) +- (0.0, 1.7211033170478025e-08)
(0.475, 1.3578341144475553e-07) +- (0.0, 1.7180926579825277e-08)
(0.45, 1.1445784344927687e-07) +- (0.0, 1.4190429371760798e-08)
(0.425, 9.588147072015695e-08) +- (0.0, 6.5564844935952024e-09)
(0.4, 8.26860087951328e-08) +- (0.0, 4.185632045375812e-09)
(0.375, 7.243463233838155e-08) +- (0.0, 3.1935915352955035e-09)
(0.35, 6.541343868549355e-08) +- (0.0, 2.493047375851807e-09)
        };
        \addplot+[error bars/.cd,y dir=both,y explicit]
        coordinates {
(0.55, 8.518653347078196e-08) +- (0.0, 1.2079749361120356e-08)
(0.525, 7.466588123202065e-08) +- (0.0, 9.061596975506167e-09)
(0.5, 6.789367263546762e-08) +- (0.0, 6.8701686532457364e-09)
(0.475, 6.290764880218925e-08) +- (0.0, 5.079235916276956e-09)
(0.45, 5.961898604780997e-08) +- (0.0, 4.364085371183952e-09)
(0.425, 5.534022926627761e-08) +- (0.0, 3.4210174226928947e-09)
(0.4, 5.177023697452459e-08) +- (0.0, 2.9426566297518416e-09)
(0.375, 4.8997101642761596e-08) +- (0.0, 2.6535373612803674e-09)
(0.35, 4.6865137483714546e-08) +- (0.0, 2.335062230590948e-09)
        };

    \nextgroupplot[
        x tick label style={rotate=45,anchor=north east,/pgf/number format/.cd, fixed, fixed zerofill,
        precision=3},
        xlabel=\normalsize{Similarity threshold $\tau$},
        ybar,
        bar width = 4pt,
        ytick = {0,2,4,6,8,10},
        ymajorgrids=true,
        ymin=0, ymax=10,
        x dir=reverse,
        xtick={0.55,0.525,0.5,0.475,0.45,0.425,0.4,0.375,0.35},
        ]
        \addplot+[error bars/.cd,y dir=both,y explicit]
        coordinates {
(0.55, 1.1038695744021594) +- (0.0, 0.04435599221662197)
(0.525, 1.375083205375272) +- (0.0, 0.048326904595002336)
(0.5, 1.6795629369039748) +- (0.0, 0.050521161696334256)
(0.475, 2.0300928911716247) +- (0.0, 0.051577435077726365)
(0.45, 2.416949548443516) +- (0.0, 0.05163791568432423)
(0.425, 2.814799364304996) +- (0.0, 0.05090654278610328)
(0.4, 3.246336146270806) +- (0.0, 0.048261275207125316)
(0.375, 3.675212513701333) +- (0.0, 0.0466812543474827)
(0.35, 4.118405724262011) +- (0.0, 0.044310477431952815)
    }; 
        \addplot+[error bars/.cd,y dir=both,y explicit] 
        coordinates {
(0.55, 1.894589221650243) +- (0.0, 0.047494155576998334)
(0.525, 2.1922578887102047) +- (0.0, 0.04774958940857826)
(0.5, 2.502533976274927) +- (0.0, 0.047703402595188925)
(0.475, 2.8421733021424407) +- (0.0, 0.04743836570960746)
(0.45, 3.201307709600951) +- (0.0, 0.04656743082332271)
(0.425, 3.5672318065314026) +- (0.0, 0.045195931397657485)
(0.4, 3.9398699152925505) +- (0.0, 0.04333215536879429)
(0.375, 4.3307179210251245) +- (0.0, 0.04109993790412969)
(0.35, 4.723944741285959) +- (0.0, 0.03874429726346105)
        };
    \nextgroupplot[
        x tick label style={rotate=45,anchor=north east,/pgf/number format/.cd, fixed, fixed zerofill,
        precision=3},        
        xlabel=\normalsize{Similarity threshold $\tau$},
        ybar,
        bar width = 4pt,
        ytick = {0,0.2,0.4,0.6,0.8,1},
        ymajorgrids=true,
        ymin=0, ymax=1,
        x dir=reverse,
        xtick={0.55,0.525,0.5,0.475,0.45,0.425,0.4,0.375,0.35},
        ]
        \addplot+[error bars/.cd,y dir=both,y explicit]
        coordinates {
(0.55, 0.6891524992466627) +- (0.0, 0.019815034931847535)
(0.525, 0.6901648980368246) +- (0.0, 0.018851457604615356)
(0.5, 0.6921079854468745) +- (0.0, 0.018429714470159918)
(0.475, 0.6786214813501149) +- (0.0, 0.01890515607834476)
(0.45, 0.6772151092713126) +- (0.0, 0.018814818633290882)
(0.425, 0.6983032819771408) +- (0.0, 0.018131071015068095)
(0.4, 0.7032498315299589) +- (0.0, 0.0176513424314291)
(0.375, 0.6917911021817797) +- (0.0, 0.019118605927346047)
(0.35, 0.6850275527916232) +- (0.0, 0.019451981617938624)
        };
        \addplot+[error bars/.cd,y dir=both,y explicit]
        coordinates {
(0.55, 0.5003773318505192) +- (0.0, 0.027706345734111716)
(0.525, 0.4954032654618007) +- (0.0, 0.02631115150825901)
(0.5, 0.47714973959233214) +- (0.0, 0.025307700360811574)
(0.475, 0.49070357583590046) +- (0.0, 0.024762105896645382)
(0.45, 0.46941707418275164) +- (0.0, 0.025430901077017763)
(0.425, 0.4726073441899249) +- (0.0, 0.025113907961910324)
(0.4, 0.4776190557576429) +- (0.0, 0.024716899189817967)
(0.375, 0.47092564339857945) +- (0.0, 0.025231059288350163)
(0.35, 0.47610197083532) +- (0.0, 0.02538669711742298)
        };
    \nextgroupplot[
        x tick label style={rotate=45,anchor=north east,/pgf/number format/.cd, fixed, fixed zerofill,
        precision=3},        
        xlabel=\normalsize{Similarity threshold $\tau$},
        ybar,
        bar width = 4pt,
        ytick = {0,5e-8,1e-7,15e-8,2e-7,25e-8,3e-7},
        ymajorgrids=true,
        ymin=0, ymax=3e-7,
        x dir=reverse,
        xtick={0.55,0.525,0.5,0.475,0.45,0.425,0.4,0.375,0.35},
        ]
        \addplot+[error bars/.cd,y dir=both,y explicit]
        coordinates {
(0.55, 9.789737634382047e-08) +- (0.0, 1.509619629450873e-08)
(0.525, 1.0134248153442495e-07) +- (0.0, 2.0950873193514798e-08)
(0.5, 9.084633278975924e-08) +- (0.0, 1.3435132185479629e-08)
(0.475, 9.176176451372477e-08) +- (0.0, 1.3077225453299941e-08)
(0.45, 1.211796163110648e-07) +- (0.0, 1.772495313426817e-08)
(0.425, 1.1469780648698859e-07) +- (0.0, 1.24952258508527e-08)
(0.4, 1.2449392599418803e-07) +- (0.0, 1.1429999559359636e-08)
(0.375, 1.1149821914416883e-07) +- (0.0, 1.0778709857706494e-08)
(0.35, 1.0421468229921368e-07) +- (0.0, 9.189024380939095e-09)
        };
        \addplot+[error bars/.cd,y dir=both,y explicit]
        coordinates {
(0.55, 1.0296220900311546e-07) +- (0.0, 1.5796023897879536e-08)
(0.525, 1.0469819520574976e-07) +- (0.0, 1.4428500514534378e-08)
(0.5, 1.0892453701762585e-07) +- (0.0, 1.3804449830618044e-08)
(0.475, 1.0120871374186033e-07) +- (0.0, 1.1255650420271346e-08)
(0.45, 8.01693774283767e-08) +- (0.0, 1.2209213290669057e-08)
(0.425, 8.243884852870267e-08) +- (0.0, 1.0328547247670333e-08)
(0.4, 8.09351576744679e-08) +- (0.0, 7.864584623603992e-09)
(0.375, 1.0392244194421314e-07) +- (0.0, 1.713762130420022e-08)
(0.35, 9.284172984298113e-08) +- (0.0, 1.3690065038929616e-08)
        };
\end{groupplot}

\definecolor{bananamania}{rgb}{0.98, 0.91, 0.71}
\definecolor{celadon}{rgb}{0.67, 0.88, 0.69}
\node at (8.7cm, 5cm) {\ref*{grouplegend}}; 
\draw[fill=bananamania,draw=none] (-0.9,4.4) rectangle (2,5.4) node[text width=3cm,align=center] at (0.5,4.9){\larger{Published \linebreak writing}};
\draw (-0.7,1.7)  node[rotate=90]{\larger Word2Vec};
\draw (-0.7,-2.3)  node[rotate=90]{\larger RoBERTa};

\draw (0.35,1.7)  node{\scriptsize{***}};
\draw (0.81,1.9)  node{\scriptsize{***}};
\draw (1.27,2.05)  node{\scriptsize{***}};
\draw (1.73,2.20)  node{\scriptsize{***}};
\draw (2.19,2.35)  node{\scriptsize{***}};
\draw (2.65,2.5)  node{\scriptsize{***}};
\draw (3.11,2.65)  node{\scriptsize{***}};
\draw (3.57,2.8)  node{\scriptsize{***}};
\draw (4.05,2.95)  node{\scriptsize{***}};

\draw (0.35,-3.2)  node{\scriptsize{***}};
\draw (0.81,-3.09)  node{\scriptsize{***}};
\draw (1.27,-2.98)  node{\scriptsize{***}};
\draw (1.73,-2.87)  node{\scriptsize{***}};
\draw (2.19,-2.76)  node{\scriptsize{***}};
\draw (2.65,-2.65)  node{\scriptsize{***}};
\draw (3.11,-2.54)  node{\scriptsize{***}};
\draw (3.57,-2.43)  node{\scriptsize{***}};
\draw (4.05,-2.32)  node{\scriptsize{***}};

\draw (5.58,2.9)  node{\scriptsize{*}};
\draw (2.19+5.23,2.95)  node{\scriptsize{*}};
\draw (2.19+5.23+0.46,3)  node{\scriptsize{**}};
\draw (2.19+5.23+0.46*2,2.99)  node{\scriptsize{***}};
\draw (2.19+5.23+0.46*3,2.99)  node{\scriptsize{***}};
\draw (2.19+5.23+0.46*4,3)  node{\scriptsize{***}};

\draw (5.58,-1.5)  node{\scriptsize{***}};
\draw (2.19+5.23-0.46*3,-1.55)  node{\scriptsize{***}};
\draw (2.19+5.23-0.46*2,-1.5)  node{\scriptsize{***}};
\draw (2.19+5.23-0.46,-1.55)  node{\scriptsize{***}};
\draw (2.19+5.23,-1.5)  node{\scriptsize{***}};
\draw (2.19+5.23+0.46,-1.55)  node{\scriptsize{***}};
\draw (2.19+5.23+0.46*2,-1.5)  node{\scriptsize{***}};
\draw (2.19+5.23+0.46*3,-1.5)  node{\scriptsize{***}};
\draw (2.19+5.23+0.46*4,-1.55)  node{\scriptsize{***}};

\draw (10.81,3)  node{\scriptsize{***}};
\draw (10.81+0.46,2.3)  node{\scriptsize{***}};
\draw (10.81+0.46*2,2)  node{\scriptsize{***}};
\draw (10.81+0.46*3,1.8)  node{\scriptsize{***}};
\draw (10.81+0.46*4,1.6)  node{\scriptsize{***}};
\draw (10.81+0.46*5,1.3)  node{\scriptsize{***}};
\draw (10.81+0.46*6,1.15)  node{\scriptsize{***}};
\draw (10.81+0.46*7,1)  node{\scriptsize{***}};
\draw (10.81+0.46*8,0.9)  node{\scriptsize{***}};

\draw (10.81+0.46*6,-2.45)  node{\scriptsize{*}};

\end{tikzpicture}
\caption{Experimental comparison between the neighbourhoods of neologisms (blue bars) and control words (red bars) in the \textbf{published writing} domain. The three plots in each row correspond to three measures: the number of \Historical{} neighbours a word has (left), how monotonically these neighbours grow in frequency (centre), and the linear regression slope of their growth (right). The x-axis on all plots corresponds to the neighbourhood size (defined by the cosine similarity threshold $\tau$). The top and bottom rows show the results with the static Word2Vec embeddings and the contextual RoBERTa embeddings respectively. Error bars represent standard error over words. The number of asterisks above a pair of bars indicates the statistical significance of their difference per Wilcoxon signed-rank test: *** for $p < 0.001$, ** for $0.001 \leq p < 0.01$, * for $0.01 \leq p < 0.05$, none for $p \geq 0.05$.}
\label{fig:coha_figure}
\end{figure*}
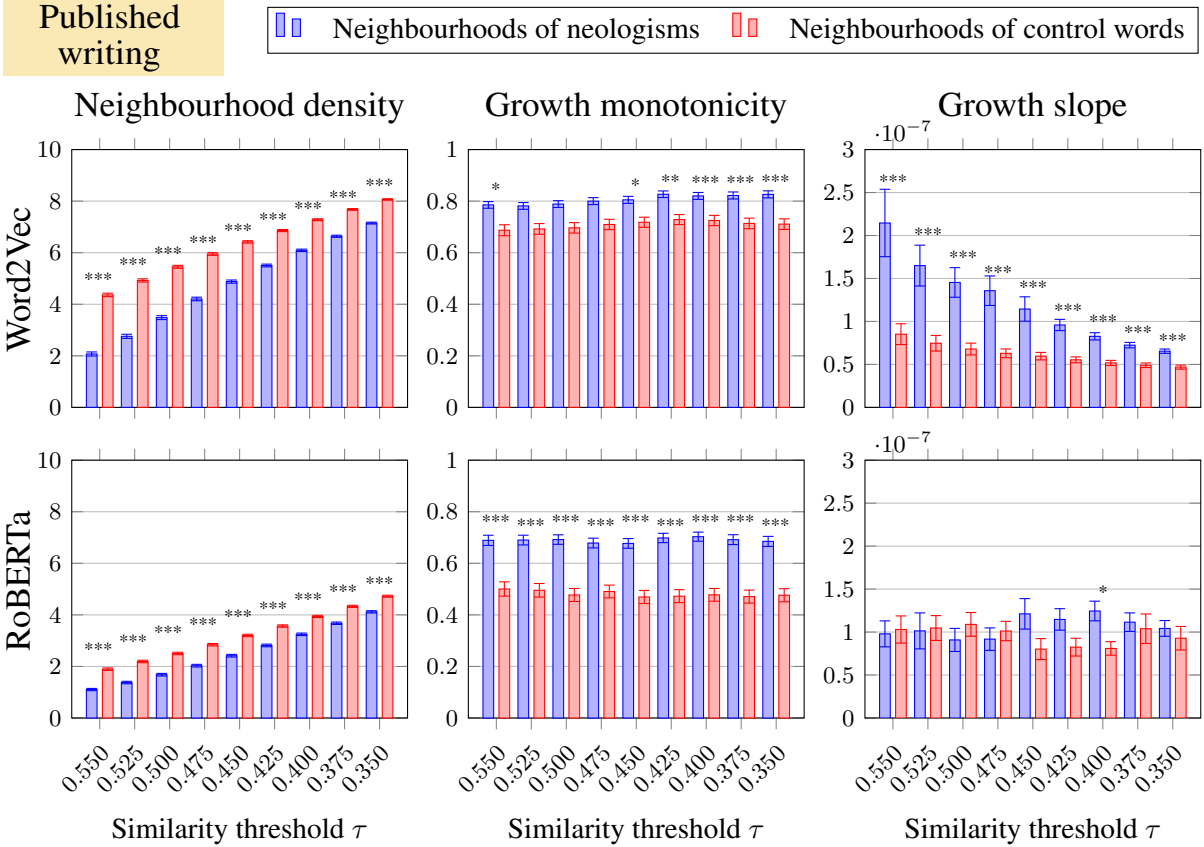
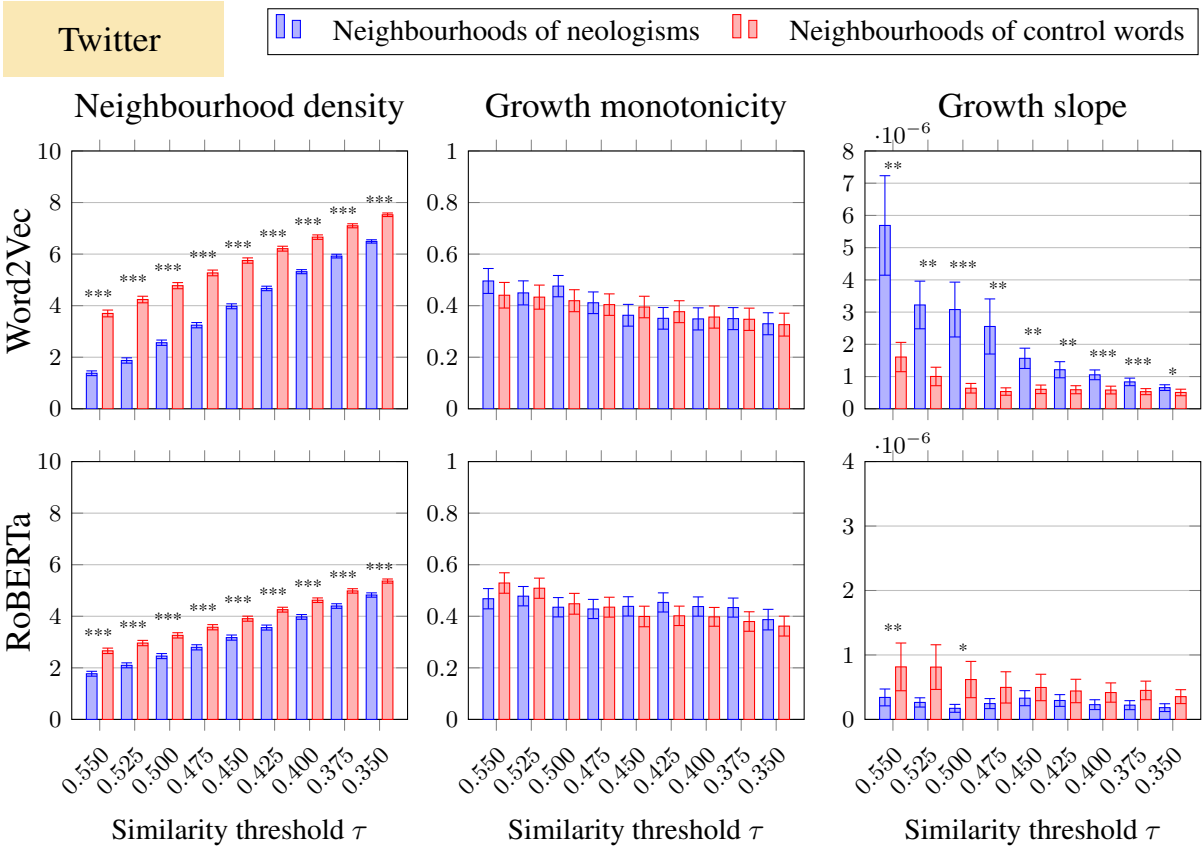
\begin{figure*}[t]
\centering
\begin{tikzpicture}
\pgfplotsset{small, every tick label/.append style={font=\footnotesize}}
\begin{groupplot}[group style = {group size = 3 by 2, horizontal sep = 0.8cm, vertical sep = 0.7cm}, width = 6cm, height = 5cm, ]
    \nextgroupplot[title = \larger{Neighbourhood density},
        legend style = { column sep = 10pt, legend columns = -1, legend to name = grouplegend4},
        x tick label style={rotate=45,anchor=north east,/pgf/number format/.cd, fixed, fixed zerofill,
        precision=3},
        ybar,
        bar width = 4pt,
        ytick = {0,2,4,6,8,10},
        ymajorgrids=true,
        ymin=0, ymax=10,
        x dir=reverse,
        xtick={0.55,0.525,0.5,0.475,0.45,0.425,0.4,0.375,0.35},
        xticklabel=\empty
        ]
        \addplot+[error bars/.cd,y dir=both,y explicit]
        coordinates {
(0.55, 1.3782154237229491) +- (0.0, 0.09192119781670546)
(0.525, 1.8737117003407244) +- (0.0, 0.10279857285562174)
(0.5, 2.5627036265675276) +- (0.0, 0.10349260521608813)
(0.475, 3.2426059378074323) +- (0.0, 0.10213242537742445)
(0.45, 3.9759681489653134) +- (0.0, 0.0954194440968001)
(0.425, 4.668142651373299) +- (0.0, 0.0869689099840585)
(0.4, 5.31816962952742) +- (0.0, 0.07900703639279315)
(0.375, 5.922490862581092) +- (0.0, 0.07196468123747936)
(0.35, 6.491668265303701) +- (0.0, 0.06469353420766294)
    }; 
    \addlegendentry{Neighbourhoods of neologisms}%
        \addplot+[error bars/.cd,y dir=both,y explicit] 
        coordinates {
(0.55, 3.6977712702863195) +- (0.0, 0.13175785371581128)
(0.525, 4.237993475414724) +- (0.0, 0.12755623053659368)
(0.5, 4.775498013175345) +- (0.0, 0.1189559366496357)
(0.475, 5.269786419253287) +- (0.0, 0.11089852845240765)
(0.45, 5.750387367006715) +- (0.0, 0.10258880471616713)
(0.425, 6.210514101122642) +- (0.0, 0.09474161813130803)
(0.4, 6.65815631768739) +- (0.0, 0.0868057153976252)
(0.375, 7.098188975526409) +- (0.0, 0.07828362620105248)
(0.35, 7.525686444805158) +- (0.0, 0.0694573545794466)
        }; \addlegendentry{Neighbourhoods of control words}
    \nextgroupplot[title=\larger{Growth monotonicity},
        x tick label style={rotate=45,anchor=north east,/pgf/number format/.cd, fixed, fixed zerofill,
        precision=3},        
        ybar,
        bar width = 4pt,
        ytick = {0,0.2,0.4,0.6,0.8,1},
        ymajorgrids=true,
        ymin=0, ymax=1,
        x dir=reverse,
        xtick={0.55,0.525,0.5,0.475,0.45,0.425,0.4,0.375,0.35},
        xticklabel=\empty
        ]
        \addplot+[error bars/.cd,y dir=both,y explicit]
        coordinates {
(0.55, 0.4957713591600491) +- (0.0, 0.04833733405274829)
(0.525, 0.44951556700541256) +- (0.0, 0.04666460103377134)
(0.5, 0.47577905255686725) +- (0.0, 0.04131299934601577)
(0.475, 0.41123885175150204) +- (0.0, 0.04198621435852517)
(0.45, 0.3627543566027481) +- (0.0, 0.04220203112147901)
(0.425, 0.35082590080672554) +- (0.0, 0.041997193284465745)
(0.4, 0.3486955986928594) +- (0.0, 0.04289825497113278)
(0.375, 0.34978354978354975) +- (0.0, 0.04271456376684658)
(0.35, 0.32987012987012987) +- (0.0, 0.04263658891782108)
        };
        \addplot+[error bars/.cd,y dir=both,y explicit]
        coordinates {
(0.55, 0.44044995993439007) +- (0.0, 0.04967056941775244)
(0.525, 0.4330738697624092) +- (0.0, 0.04670275648749559)
(0.5, 0.4192153875403507) +- (0.0, 0.04261842247185961)
(0.475, 0.40402406545336783) +- (0.0, 0.041678924951841216)
(0.45, 0.39475982532751086) +- (0.0, 0.04175571847181587)
(0.425, 0.37662337662337664) +- (0.0, 0.042470695210921235)
(0.4, 0.35584415584415585) +- (0.0, 0.042963268837683796)
(0.375, 0.3471861471861472) +- (0.0, 0.04320050970842562)
(0.35, 0.3264069264069264) +- (0.0, 0.04435982894504708)
        };
    \nextgroupplot[title=\larger{Growth slope},
        x tick label style={rotate=45,anchor=north east,/pgf/number format/.cd, fixed, fixed zerofill,
        precision=3},        
        ybar,
        bar width = 4pt,
        ytick = {0,1e-6,2e-6,3e-6,4e-6,5e-6,6e-6,7e-6,8e-6},
        ymajorgrids=true,
        ymin=0, ymax=0.000008,
        x dir=reverse,
        xtick={0.55,0.525,0.5,0.475,0.45,0.425,0.4,0.375,0.35},
        xticklabel=\empty
        ]
        \addplot+[error bars/.cd,y dir=both,y explicit]
        coordinates {
(0.55, 5.688792591318284e-06) +- (0.0, 1.5438118306020264e-06)
(0.525, 3.2215508173774485e-06) +- (0.0, 7.391101425692026e-07)
(0.5, 3.0796318602860956e-06) +- (0.0, 8.507000635154221e-07)
(0.475, 2.5541565073887944e-06) +- (0.0, 8.54863209458939e-07)
(0.45, 1.5659721333340562e-06) +- (0.0, 3.1472280012961816e-07)
(0.425, 1.212216289569127e-06) +- (0.0, 2.4980860453021097e-07)
(0.4, 1.0541954107707827e-06) +- (0.0, 1.5169522629338422e-07)
(0.375, 8.346447439473408e-07) +- (0.0, 1.1717669883725811e-07)
(0.35, 6.569254207555663e-07) +- (0.0, 8.879016030853196e-08)
        };
        \addplot+[error bars/.cd,y dir=both,y explicit]
        coordinates {
(0.55, 1.6060586893906323e-06) +- (0.0, 4.548164475931179e-07)
(0.525, 1.0029125835210578e-06) +- (0.0, 2.867633720237861e-07)
(0.5, 6.377892812100042e-07) +- (0.0, 1.4905360019635077e-07)
(0.475, 5.350183376640548e-07) +- (0.0, 1.1512168609852713e-07)
(0.45, 6.040084049532101e-07) +- (0.0, 1.3265992289689212e-07)
(0.425, 5.912559493846835e-07) +- (0.0, 1.2485999075254826e-07)
(0.4, 5.817093703436326e-07) +- (0.0, 1.203727782746463e-07)
(0.375, 5.32152977996068e-07) +- (0.0, 9.352763202339896e-08)
(0.35, 5.068908186653528e-07) +- (0.0, 1.0122104418361999e-07)
        };

    \nextgroupplot[
        x tick label style={rotate=45,anchor=north east,/pgf/number format/.cd, fixed, fixed zerofill,
        precision=3},
        xlabel=\normalsize{Similarity threshold $\tau$},
        ybar,
        bar width = 4pt,
        ytick = {0,2,4,6,8,10},
        ymajorgrids=true,
        ymin=0, ymax=10,
        x dir=reverse,
        xtick={0.55,0.525,0.5,0.475,0.45,0.425,0.4,0.375,0.35},
        ]
        \addplot+[error bars/.cd,y dir=both,y explicit]
        coordinates {
(0.55, 1.7712475846545255) +- (0.0, 0.09498341395195437)
(0.525, 2.0972330705886373) +- (0.0, 0.09642658868584476)
(0.5, 2.4558445859296567) +- (0.0, 0.09760656483613304)
(0.475, 2.795443077662503) +- (0.0, 0.10022020729768383)
(0.45, 3.1730768594218652) +- (0.0, 0.10000469076897028)
(0.425, 3.56219421181132) +- (0.0, 0.09678748769050796)
(0.4, 3.975443713086906) +- (0.0, 0.09236819963906674)
(0.375, 4.397794273802948) +- (0.0, 0.08820023611447364)
(0.35, 4.824577139885859) +- (0.0, 0.08383309623334642)
    }; 
        \addplot+[error bars/.cd,y dir=both,y explicit] 
        coordinates {
(0.55, 2.6602122734770215) +- (0.0, 0.10559297643803399)
(0.525, 2.961656624329766) +- (0.0, 0.10351811066489988)
(0.5, 3.2598344143150038) +- (0.0, 0.10251556955604876)
(0.475, 3.57475848411932) +- (0.0, 0.10111037430679766)
(0.45, 3.9067284796253983) +- (0.0, 0.09818079860610639)
(0.425, 4.256401449950622) +- (0.0, 0.09395366847334904)
(0.4, 4.6227765905387646) +- (0.0, 0.08885094975005649)
(0.375, 4.983763628337374) +- (0.0, 0.08536495862019963)
(0.35, 5.365827774674199) +- (0.0, 0.08051961930243832)
        };
    \nextgroupplot[
        x tick label style={rotate=45,anchor=north east,/pgf/number format/.cd, fixed, fixed zerofill,
        precision=3},        
        xlabel=\normalsize{Similarity threshold $\tau$},
        ybar,
        bar width = 4pt,
        ytick = {0,0.2,0.4,0.6,0.8,1},
        ymajorgrids=true,
        ymin=0, ymax=1,
        x dir=reverse,
        xtick={0.55,0.525,0.5,0.475,0.45,0.425,0.4,0.375,0.35},
        ]
        \addplot+[error bars/.cd,y dir=both,y explicit]
        coordinates {
(0.55, 0.46808925000824436) +- (0.0, 0.03924301541610456)
(0.525, 0.4779549436916483) +- (0.0, 0.037528747608726154)
(0.5, 0.4350954696804588) +- (0.0, 0.03742150294444425)
(0.475, 0.4283045744819057) +- (0.0, 0.037163588862329185)
(0.45, 0.43836849082470725) +- (0.0, 0.037418468204692835)
(0.425, 0.45373352001604417) +- (0.0, 0.037145096651716024)
(0.4, 0.43776760037140494) +- (0.0, 0.03733253497723201)
(0.375, 0.4337662337662338) +- (0.0, 0.036898542197134725)
(0.35, 0.387012987012987) +- (0.0, 0.039875828102679436)
        };
        \addplot+[error bars/.cd,y dir=both,y explicit]
        coordinates {
(0.55, 0.529031703707474) +- (0.0, 0.03968207875555747)
(0.525, 0.5088190556674965) +- (0.0, 0.03912035621033174)
(0.5, 0.448497545103277) +- (0.0, 0.04035444619788135)
(0.475, 0.4353526151324728) +- (0.0, 0.038511343307960834)
(0.45, 0.3992271117076168) +- (0.0, 0.03987327928714583)
(0.425, 0.40173891274097817) +- (0.0, 0.03764010694415259)
(0.4, 0.39781463737435224) +- (0.0, 0.036487357927503974)
(0.375, 0.37964227963680097) +- (0.0, 0.03785774134138663)
(0.35, 0.3619047619047619) +- (0.0, 0.038508021487640244)
        };
    \nextgroupplot[
        x tick label style={rotate=45,anchor=north east,/pgf/number format/.cd, fixed, fixed zerofill,
        precision=3},        
        xlabel=\normalsize{Similarity threshold $\tau$},
        ybar,
        bar width = 4pt,
        ytick = {0,1e-6,2e-6,3e-6,4e-6},
        ymajorgrids=true,
        ymin=0, ymax=4e-6,
        x dir=reverse,
        xtick={0.55,0.525,0.5,0.475,0.45,0.425,0.4,0.375,0.35},
        ]
        \addplot+[error bars/.cd,y dir=both,y explicit]
        coordinates {
(0.55, 3.4107048684424106e-07) +- (0.0, 1.306593886376372e-07)
(0.525, 2.6358536046808184e-07) +- (0.0, 7.185879268599472e-08)
(0.5, 1.7176475727235526e-07) +- (0.0, 6.124054816125138e-08)
(0.475, 2.4546851130963e-07) +- (0.0, 7.780704971852975e-08)
(0.45, 3.293817965878696e-07) +- (0.0, 1.1675335449066534e-07)
(0.425, 2.9338538024147997e-07) +- (0.0, 9.12351343970831e-08)
(0.4, 2.2895659931599478e-07) +- (0.0, 7.632673298756027e-08)
(0.375, 2.2129716051733744e-07) +- (0.0, 6.981014471571439e-08)
(0.35, 1.8255530223716186e-07) +- (0.0, 6.060491489889317e-08)
        };
        \addplot+[error bars/.cd,y dir=both,y explicit]
        coordinates {
(0.55, 8.149501816644447e-07) +- (0.0, 3.7026718886858926e-07)
(0.525, 8.116865074036755e-07) +- (0.0, 3.4643220265575624e-07)
(0.5, 6.185935854129184e-07) +- (0.0, 2.814706273741946e-07)
(0.475, 4.966687025831565e-07) +- (0.0, 2.4246010337526584e-07)
(0.45, 4.954951076477607e-07) +- (0.0, 2.0512356459211643e-07)
(0.425, 4.410571989904686e-07) +- (0.0, 1.817369446239991e-07)
(0.4, 4.1671454436862175e-07) +- (0.0, 1.497203096434978e-07)
(0.375, 4.4974469998904745e-07) +- (0.0, 1.4407491728429056e-07)
(0.35, 3.53297239190844e-07) +- (0.0, 1.0792454206684203e-07)
        };
\end{groupplot}

\definecolor{bananamania}{rgb}{0.98, 0.91, 0.71}
\node at (8.7cm, 5cm) {\ref*{grouplegend4}}; 
\draw[fill=bananamania,draw=none] (-0.9,4.4) rectangle (2,5.4) node[text width=3cm,align=center] at (0.5,4.9){\larger{Twitter}};
\draw (-0.7,1.7)  node[rotate=90]{\larger Word2Vec};
\draw (-0.7,-2.3)  node[rotate=90]{\larger RoBERTa};

\draw (0.35,1.5)  node{\scriptsize{***}};
\draw (0.81,1.7)  node{\scriptsize{***}};
\draw (1.27,2.05-0.2)  node{\scriptsize{***}};
\draw (1.73,2.20-0.18)  node{\scriptsize{***}};
\draw (2.19,2.35-0.2)  node{\scriptsize{***}};
\draw (2.65,2.5-0.2)  node{\scriptsize{***}};
\draw (3.11,2.65-0.2)  node{\scriptsize{***}};
\draw (3.57,2.8-0.2)  node{\scriptsize{***}};
\draw (4.05,2.95-0.2)  node{\scriptsize{***}};

\draw (0.35,-3.2+0.2)  node{\scriptsize{***}};
\draw (0.81,-3.09+0.2)  node{\scriptsize{***}};
\draw (1.27,-2.98+0.2)  node{\scriptsize{***}};
\draw (1.73,-2.87+0.2)  node{\scriptsize{***}};
\draw (2.19,-2.76+0.2)  node{\scriptsize{***}};
\draw (2.65,-2.65+0.2)  node{\scriptsize{***}};
\draw (3.11,-2.54+0.2)  node{\scriptsize{***}};
\draw (3.57,-2.43+0.2)  node{\scriptsize{***}};
\draw (4.05,-2.32+0.2)  node{\scriptsize{***}};

\draw (10.81,3.2)  node{\scriptsize{**}};
\draw (10.81+0.46,1.9)  node{\scriptsize{**}};
\draw (10.81+0.46*2,1.88)  node{\scriptsize{***}};
\draw (10.81+0.46*3,1.6)  node{\scriptsize{**}};
\draw (10.81+0.46*4,1)  node{\scriptsize{**}};
\draw (10.81+0.46*5,0.85)  node{\scriptsize{**}};
\draw (10.81+0.46*6,0.7)  node{\scriptsize{***}};
\draw (10.81+0.46*7,0.6)  node{\scriptsize{***}};
\draw (10.81+0.46*8,0.47)  node{\scriptsize{*}};

\draw (10.81,-2.93)  node{\scriptsize{**}};
\draw (10.81+0.46*2,-3.2)  node{\scriptsize{*}};

\end{tikzpicture}
\caption{Experimental comparison between the neighbourhoods of neologisms (blue bars) and control words (red bars) in the \textbf{Twitter} domain. The three plots in each row correspond to three measures: the number of \Historical{} neighbours a word has (left), how monotonically these neighbours grow in frequency (centre), and the linear regression slope of their growth (right). The x-axis on all plots corresponds to the neighbourhood size (defined by the cosine similarity threshold $\tau$). The top and bottom rows show the results with the static Word2Vec embeddings and the contextual RoBERTa embeddings respectively. Error bars represent standard error over words. The number of asterisks above a pair of bars indicates the statistical significance of their difference per Wilcoxon signed-rank test: *** for $p < 0.001$, ** for $0.001 \leq p < 0.01$, * for $0.01 \leq p < 0.05$, none for $p \geq 0.05$.
\label{fig:twitter_figure}}
\end{figure*}

Example neologism--control pairs for either data domain are shown in \Tref{tab:pair-examples}. We find that the cosine similarity constraint often adds an extra semantic or morphological connection to the resulting word pairs. It can manifest as conceptual similarity \WordForm{(e-mail:message)}, morphological overlap \WordForm{(baecation:staycation)}, matching part of speech and form \WordForm{(cringiest:silliest)}, or creative spelling choice \WordForm{(bruhhhhh:niceeeee)}.

\subsection{Experimental setup \label{sec:setup}}

We operationalize the supply and demand hypotheses following our prior work \cite{ryskina-etal-2020-new}, with minor modifications to make the analysis more robust (detailed comparison in Appendix \ref{sec:appendix-diff}).
Let $N_{\tau}(w)$ denote a neighbourhood of a word $w$, i.e., the set of words in the \Historical{} embedding space whose cosine similarity to $w$ exceeds a given threshold $\tau$. If $w$ is a neologism, which may not be present in the \Historical{} vocabulary $V_H$, we use its \Modern{} embedding (Procrustes-projected for Word2Vec or as-is for RoBERTa).

To quantify the supply hypothesis, we measure a function of the neighbourhood density:
\begin{equation}
    d(w, \tau) = \log (|N_{\tau}(w)| + 1).
\label{eq:density}
\end{equation}

The demand hypothesis measures how the neighbourhood words change in popularity. At each timestep $t \in \{1, \ldots, T\}$, we measure how much of the corpus do the words in $N_{\tau}(w)$ account for:
\begin{equation}
 p_{\tau}(w, t) = \frac{\sum_{u \in N_{\tau}(w)} c_t(u)}{\sum_{u \in V_H} c_t(u)} 
\label{eq:growth}
\end{equation}
We measure the monotonicity of this time series (Spearman correlation $\rho$ with the timesteps):
\begin{equation}
   m(w, \tau) = \rho \left(\{1, \ldots, T\}, 
\{p_{\tau}(w, t) \}_{t=1}^{T}\right) 
\label{eq:growth-monotonicity}
\end{equation}
and the linear regression slope of its growth:
\begin{multline}
    r(w, \tau) = 
        \frac{1}{d(w, \tau)} \times \\
    \frac{\sum_{t=1}^T (t - \bar{t})(p_\tau(w, t) - \bar{p}_\tau(w))}{\sum_{t=1}^T (t - \bar{t})^2}, 
\label{eq:growth-slope}
\end{multline}
where $\bar{p}_\tau(w) = \frac{1}{T}[\sum_{t=1}^T p_\tau(w, t)]$ and $\bar{t} = \frac{(1+T)}{2}$.

In the experiments described in the next section, we compute the mean values of these three metrics for the neighbourhoods of neologisms and the neighbourhoods of control words over a range of neighbourhood sizes defined by the threshold $\tau$. 

\section{Results \label{sec:results}}

\begin{table*}[thb]
    \centering
    \begin{tabular}{llp{5.5cm}}
        \toprule
        \multirow{2}{3.5cm}{Neologism formation mechanism} & \multicolumn{1}{c}{Published writing (1810--2012)} & \multicolumn{1}{c}{Twitter (2007--2021)} \\
        & \multicolumn{1}{c}{\% neologisms (examples)} & \multicolumn{1}{c}{\% neologisms (examples)} \\
        \midrule
        Abbreviation & \hphantom{<}6\% \WordForm{(blog, condo, postdoc)} & 20\% \WordForm{(bae, afab, incel)} \\
        Blend & \hphantom{<}4\% \WordForm{(pixel, camcorder)} & 14\% \WordForm{(presstitutes, weeksary)} \\
        Borrowing & 14\% \WordForm{(falafel, video)} & 11\% \WordForm{(mukbang, dololo)} \\ 
        Compound & 45\% \WordForm{(laptop, cyberpunk)} & 19\% \WordForm{(deepfake, headcanon)} \\ 
        POS conversion & \hphantom{<}7\% \WordForm{(startup, aerobics)} & \hphantom{0}8\% \WordForm{(snapchatting, mutuals)} \\
        Derivation & 29\% \WordForm{(wellness, facilitator)} & 20\% \WordForm{(anti-mask, unalive)} \\
        Sense & \hphantom{<}4\% \WordForm{(analog, firewall)} & \hphantom{0}4\% \WordForm{(catfishing)} \\
        Spelling & <1\% \WordForm{(byte, gangsta)} & 13\% \WordForm{(bæ, 5ever, sksksk)} \\

        \bottomrule
    \end{tabular}
    \caption{Percentage of neologisms in each domain
    by their formation mechanism. Our manual categorization draws on the Oxford English Dictionary where available, and allows for the same word to fall into multiple categories. 
    Category definitions (details in \Sref{sec:appendix-word-formation}): abbreviation = clipping, shortening, acronym; blend = combining parts of existing words; borrowing = loanwords from other languages; compound = combining full existing words; POS conversion = change in part of speech; derivation = morphological derivation, e.g., affixation; sense = widely used existing words acquiring new senses; spelling = any creative spellings, including alterations of existing words.
    \label{tab:formation}}
\end{table*}

\subsection{Published writing \label{sec:results-cocacoha}}

\Fref{fig:coha_figure} shows how the mean neighbourhood density (left) and the mean neighbour frequency growth rate (middle, right) differ between neologisms and controls in the published writing corpus over a range of neighbourhood sizes. The top row of charts shows the results for the static embeddings experiment, which we perform to ensure that our original findings \cite{ryskina-etal-2020-new} still hold given the changes in the neologism--control pairing criteria (\Sref{sec:control}) and the operationalization of the hypotheses (\Sref{sec:setup}). We successfully reproduce the original results, finding support for both the supply and the demand hypotheses: neologisms have fewer neighbours than the control words (\Fref{fig:coha_figure}, top left) but their neighbours grow in frequency faster (\Fref{fig:coha_figure}, top centre \& right).

The bottom row of charts shows the results with the RoBERTa embeddings: we confirm that the trends remain the same when we perform the experiment using contextual embeddings. However, the significance of the findings for  the demand hypothesis is less conclusive: the difference in neighbour frequency growth between neologism and control neighbourhoods is less pronounced for the regression slope metric (\Fref{fig:coha_figure}, bottom right) than for the monotonicity metric  (\Fref{fig:coha_figure}, bottom centre), unlike the static embedding experiment where both metrics show a significant difference for most neighbourhood sizes.

\subsection{Twitter \label{sec:results-twitter}}

The results for our Twitter corpus are presented in 
\Fref{fig:twitter_figure}. As in the published writing corpus, we find significant evidence for the supply hypothesis (left column),
suggesting that neology on social media is also partly driven by the pressure to fill in gaps in the lexicon. However, the evidence for the demand hypothesis is weaker: the frequency growth monotonicity metric (middle column) does not show a significant difference between neologisms and controls, 
and the frequency growth slope (right column) is higher for neighbours of neologism than for neighbours of control words only with Word2Vec embeddings (top right).
This suggests that the frequency growth is less important in Twitter neology, though we discuss some alternative explanations in the following section.

\section{Discussion}
\label{sec:discussion}

\paragraph{Domain and corpus differences} 
The informal and participatory nature of social media encourages creativity \cite{peppler2013social}. With language in particular, additional constraints like automated moderation \cite{ungless2025lesbean} or typing and text rendering issues \cite{ryskina-etal-2020-phonetic-2} further incentivize users to experiment with spelling and word creation. These domain-specific pressures are reflected in how social media neologisms are formed and in what neighbourhoods they emerge.

We categorize all verified neologisms by the mechanism of their formation, using dictionaries and other resources (\Tref{tab:formation}; details in Appendix \ref{sec:appendix-word-formation}). Vast majority of the neologisms in published writing are created by recombining existing morphemes, either through compounding (e.g., \emph{airfare}) or morphological derivation (e.g., \emph{interconnectedness}). As shown in \Tref{tab:formation}, Twitter neologisms are much more diverse and creative. Blends and portmanteaus are more prominent, especially in fandom culture (e.g., \emph{Bettie + Archie $\rightarrow$ Barchie}). Expressions of orthographic creativity are common, including phonetic respellings of existing words and phrases (e.g., \emph{on that $\rightarrow$ onnat}), expressing intonation or emotion (e.g., \emph{stop} $\rightarrow$ \emph{stahp}, \emph{lit} $\rightarrow$ \emph{littttt}), puns and wordplay (e.g., \emph{forever} $\rightarrow$ \emph{4ever} $\rightarrow$ \emph{5ever}, \emph{democrats} $\rightarrow$ \emph{demonrats}), and even completely novel coinages (e.g., \emph{sksksk} to express laughter). Twitter's character limit also encourages abbreviation, such as clipping (\emph{cryptocurrency} $\rightarrow$ \emph{crypto}), shortening (\emph{season} $\rightarrow$ \emph{szn}), and acronyms (\emph{one of my followers} $\rightarrow$ \emph{oomf}). Finally, our published writing corpus focuses on American English and is mostly representative of its ``standard'' variety \cite{milroy2012}, while the Twitter data features African American English \cite[known to be one of the main sources of lexical innovation on the platform;][]{grieve2018mapping}, other World Englishes, and code-switching with other languages \cite{dogruoz-etal-2021-survey}. Although the overall share of neologisms of foreign origin (see `Borrowing' in \Tref{tab:formation}) is lower for Twitter than for the published writing corpus, these words come from a more diverse set source languages and often include slang from other cultures (e.g., \emph{ahre}; \citealp{de2021marcador}). In many ways, Twitter neologisms are less typical than the published writing ones, which could explain the differences in the neighbourhood statistics between the two domains.

The differences in the corpus construction also impact the experimental results: \phistorical{} contains 18 timesteps (decades from 1800s to 1980s), while \thistorical{} has only four (years from 2007 to 2010). As a result, the estimates of the frequency growth monotonicity measure (introduced originally for the published writing corpus; \citealp{ryskina-etal-2020-new}) are too noisy for Twitter (\Fref{fig:twitter_figure}, middle column). However, our frequency growth slope measure shows support for the demand hypotheses both on Twitter (\Fref{fig:twitter_figure}, top left) and in published writing (\Fref{fig:coha_figure}, top left) for neighbourhoods in the static embedding space.

Finally, we should note that our definition of a neologism---a new form--meaning pair that spreads beyond its initial user(s) to a wider community---does not distinguish between true \emph{broadening of use} and the \emph{growth of the user community itself}. Especially for Twitter, the trends we see might be affected by how the platform's user base changes. The increased use of South African slang words in English contexts could be due to an influx of users from South Africa, and the growing popularity of terms associated with K-pop (Korean popular music) could be explained by more of the platform's users becoming interested in K-pop. Future work on social media neology could focus specifically on posts aimed at a more general audience (such as the \texttt{r/CasualConversation} subreddit), additionally evaluate the degree of a neologism's social or geographic diffusion \cite{10.1371/journal.pone.0113114,grieve2018mapping},
or jointly consider the two domains explored in this paper: a word's adoption by the published/mass media is a strong indicator of it entering the mainstream vocabulary.

\paragraph{Impact of embedding choice} The experimental results for the demand hypothesis differ depending on whether static or contextual embeddings are used. The difference is especially notable for Twitter, where the neighbours of the neologisms in the RoBERTa space seem to grow in frequency \emph{less} than the neighbours of the control words (\Fref{fig:twitter_figure}, bottom right)---the opposite of what the hypothesis predicts. After inspecting the neighbourhoods of the neologisms, we find that the distance in the RoBERTa space is substantially affected by subword token overlap, especially if any of the subwords are rare. Words sharing a root tend to be closer in this space, which is useful for neologisms created by derivation or compounding, as is common in published writing (\Tref{tab:formation}). However, Twitter abbreviations, blends, and creative spellings often contain unusual character sequences and end up being over-segmented by the RoBERTa tokenizer. For example, the nearest neighbours for \emph{smol} are other spelling variants starting with \emph{sm-} \emph{(smthin, smtimes, smoooooth)} rather than semantically similar words like \emph{cute}. This suggests that contextual embeddings may not be the best tool for studying social media neologisms specifically. In general, contextual embeddings might be better suited for a different operationalization of the hypotheses (e.g., using neighbourhoods of senses rather than words).

\section{Conclusion}

In this paper, we compare two semantic correlates of neology between two domains:  historical published writing (books and articles) and social media (Twitter). We collect a large corpus of English tweets and extend our earlier methodology \cite{ryskina-etal-2020-new} to this data in order to test two hypotheses about neology: that new form--meaning pairs emerge to fill semantic gaps (supply hypothesis) and that they emerge in topics that are rising in popularity (demand hypothesis). We operationalize the hypotheses using either static or contextual word embeddings, though we find the latter less suitable for social media data due to the impact of tokenization. For published writing, we reproduce our earlier results, finding evidence for both hypotheses. For Twitter, we find that though both hypotheses are supported, the evidence for the demand hypothesis is less conclusive. While it can be partially attributed to the differences in the corpus construction, we hypothesize that the creative tendencies of social media language use (evident in the neologism formation mechanisms prevalent on the platform) might be competing with the demand for word creation driven by the development of novel concepts in topics of growing popularity.

\section*{Limitations}

As acknowledged in \Sref{sec:discussion}, there are limitations to extending our frequency-based framing of neology from books and newspapers to social media. While social media provides a unique opportunity to study linguistic innovation in a much wider and more diverse population of language users, the composition of this population can also change rapidly, and it can be difficult to tell whether the observed frequency effects are due to the neologism's spread across communities of users or simply to the growth of the particular community it originated from. Public social media posts, despite being accessible to any user, are also not necessarily meant to be understood by a general audience, which further obscures the distinction between occasional in-group language play and more lasting and widespread language change.

We would also like to acknowledge several methodological caveats. Because of the strict neologism--control matching criteria, not all identified neologisms are matched to control words (\Sref{sec:control}) and therefore included in the final analysis, which may affect its findings. Certain differences in the design choices made for each of the two domains (the relative lengths of the \Historical{} and the \Modern{} time periods, the choice to limit neologisms to nouns only or to allow other parts of speech) might also contribute to the observed differences. Finally, pretrained contextual embeddings, while commonly used in language change studies, inevitably depend on their original pretraining corpus, which is not specifically aligned with either of our domains or time periods; future work could explore using domain- and time-specific masked language models instead.

\section*{Acknowledgments}

Resources used in preparing this research were provided, in part, by the Province of Ontario, the Government of Canada through CIFAR, and companies sponsoring the Vector Institute (\url{www.vectorinstitute.ai/partnerships/current-partners/}). We would like to thank Aria Haghighi for his comments during the conceptualization of the project and for the help with data collection, Eduard Hovy and Roger Levy for their feedback on an earlier version of this paper, Stas Kashepava for the contributions to the neologism formation analysis, and the two anonymous reviewers for their valuable suggestions.

\bibliography{custom,anthology}

@book{breal1904essai,
  title={Essai de s{\'e}mantique (science des significations).},
  author={Br{\'e}al, Michel},
  year={1904},
  publisher={Hachette}
}

@book{lenci2023distributional,
    author = {Lenci, Alessandro and Sahlgren, Magnus},
    title = {Distributional Semantics},
    publisher = {Cambridge University Press},
    year = 2023,
}

@article{davies2012coha,
  title={Expanding horizons in historical linguistics with the 400-million word {Corpus of Historical American English}},
  author={Davies, Mark},
  journal={Corpora},
  volume={7},
  number={2},
  pages={121--157},
  year={2012},
  publisher={Edinburgh University Press}
}

@article{davies2009coca,
  title={The 385+ million word {Corpus of Contemporary American English} (1990--2008+): Design, architecture, and linguistic insights},
  author={Davies, Mark},
  journal={International Journal of Corpus Linguistics},
  volume={14},
  number={2},
  pages={159--190},
  year={2009},
  publisher={John Benjamins}
}

@article{tahmasebi2021survey,
  title={Survey of computational approaches to lexical semantic change detection},
  author={Tahmasebi, Nina and Borin, Lars and Jatowt, Adam},
  journal={Computational approaches to semantic change},
  volume={6},
  number={1},
  year={2021},
  publisher={Language Science Press Berlin}
}

@inproceedings{mikolov2013distributed,
 author = {Mikolov, Tomas and Sutskever, Ilya and Chen, Kai and Corrado, Greg and Dean, Jeff},
 booktitle = {Advances in Neural Information Processing Systems},
 editor = {C.J. Burges and L. Bottou and M. Welling and Z. Ghahramani and K.Q. Weinberger},
 pages = {},
 publisher = {Curran Associates, Inc.},
 title = {Distributed Representations of Words and Phrases and their Compositionality},
 url = {https://proceedings.neurips.cc/paper_files/paper/2013/file/9aa42b31882ec039965f3c4923ce901b-Paper.pdf},
 volume = {26},
 year = {2013}
}

@article{hopcroft1973n,
  title={An $n^{5/2}$ algorithm for maximum matchings in bipartite graphs},
  author={Hopcroft, John E. and Karp, Richard M.},
  journal={SIAM Journal on computing},
  volume={2},
  number={4},
  pages={225--231},
  year={1973},
  publisher={SIAM}
}

@article{liu2019roberta,
  title={{RoBERTa}: A robustly optimized {BERT} pretraining approach},
  author={Liu, Yinhan and Ott, Myle and Goyal, Naman and Du, Jingfei and Joshi, Mandar and Chen, Danqi and Levy, Omer and Lewis, Mike and Zettlemoyer, Luke and Stoyanov, Veselin},
  journal={arXiv preprint arXiv:1907.11692},
  url={https://arxiv.org/abs/1907.11692},
  year={2019}
}

@inproceedings{kali2024cognitive,
  title={Cognitive Factors in Word Sense Decline},
  author={Kali, Aniket and Xu, Yang and Stevenson, Suzanne},
  booktitle={Proceedings of the Annual Meeting of the Cognitive Science Society},
  volume={46},
  year={2024},
  url={https://escholarship.org/uc/item/0142p2fx}
}

@article{karjus2020quantifying,
  title={Quantifying the dynamics of topical fluctuations in language},
  author={Karjus, Andres and Blythe, Richard A. and Kirby, Simon and Smith, Kenny},
  journal={Language Dynamics and Change},
  volume={10},
  number={1},
  pages={86--125},
  year={2020},
  publisher={Brill},
  url={https://arxiv.org/abs/1806.00699}
}

@article{ungless2025lesbean,
  title={Le\$bean or lesbian? {A} survey of marginalised users' motivations for obfuscation on {TikTok}},
  author={Ungless, Eddie L. and Markl, Nina and Ross, Bj{\"o}rn},
  journal={Behaviour \& Information Technology},
  url={https://www.tandfonline.com/doi/full/10.1080/0144929X.2025.2553152},
  pages={1--26},
  year={2025},
  publisher={Taylor \& Francis}
}

@inproceedings{xu2019predictability,
  title={A predictability-distinctiveness trade-off in the historical emergence of word forms},
  author={Xu, Aotao and Ramiro, Christian and Xu, Yang},
  booktitle={Proceedings of the Annual Meeting of the Cognitive Science Society},
  volume={41},
  year={2019},
  url={https://escholarship.org/uc/item/6tb249bt}
}

@inproceedings{sasse-etal-2025-making,
    title = "Making {FETCH}! Happen: Finding Emergent Dog Whistles Through Common Habitats",
    author = "Sasse, Kuleen  and
      Aguirre, Carlos Alejandro  and
      Cachola, Isabel  and
      Levy, Sharon  and
      Dredze, Mark",
    editor = "Che, Wanxiang  and
      Nabende, Joyce  and
      Shutova, Ekaterina  and
      Pilehvar, Mohammad Taher",
    booktitle = "Proceedings of the 63rd Annual Meeting of the Association for Computational Linguistics (Volume 1: Long Papers)",
    month = jul,
    year = "2025",
    address = "Vienna, Austria",
    publisher = "Association for Computational Linguistics",
    url = "https://aclanthology.org/2025.acl-long.284/",
    doi = "10.18653/v1/2025.acl-long.284",
    pages = "5687--5709",
    ISBN = "979-8-89176-251-0",
}

@article{ananthasubramaniam2024networks,
  title={Networks and identity drive the spatial diffusion of linguistic innovation in urban and rural areas},
  author={Ananthasubramaniam, Aparna and Jurgens, David and Romero, Daniel M},
  journal={npj Complexity},
  volume={1},
  number={1},
  pages={14},
  year={2024},
  publisher={Nature Publishing Group UK London},
  url={https://www.nature.com/articles/s44260-024-00009-9}
}

@inproceedings{xu2023predicting,
  title={Predicting strategy choice in word formation: A case study of reuse and compounding},
  author={Xu, Aotao and Kemp, Charles and Frermann, Lea and Xu, Yang},
  booktitle={Proceedings of the Annual Meeting of the Cognitive Science Society},
  volume={45},
  year={2023},
  url={https://escholarship.org/uc/item/1mm0w4w0}
}

@article{karjus2021conceptual,
  title={Conceptual similarity and communicative need shape colexification: An experimental study},
  author={Karjus, Andres and Blythe, Richard A and Kirby, Simon and Wang, Tianyu and Smith, Kenny},
  journal={Cognitive Science},
  volume={45},
  number={9},
  pages={e13035},
  year={2021},
  publisher={Wiley Online Library},
  url={https://onlinelibrary.wiley.com/doi/10.1111/cogs.13035}
}

@article{10.1145/3672393,
author = {Periti, Francesco and Montanelli, Stefano},
title = {Lexical Semantic Change through Large Language Models: a Survey},
year = {2024},
issue_date = {November 2024},
publisher = {Association for Computing Machinery},
address = {New York, NY, USA},
volume = {56},
number = {11},
issn = {0360-0300},
url = {https://doi.org/10.1145/3672393},
doi = {10.1145/3672393},
journal = {ACM Comput. Surv.},
month = jun,
articleno = {282},
numpages = {38}
}

@article{michel2011quantitative,
  title={Quantitative analysis of culture using millions of digitized books},
  author={Michel, Jean-Baptiste and Shen, Yuan Kui and Presser Aiden, Aviva and Veres, Adrian and Gray, Matthew K. and Google Books Team and Pickett, Joseph P. and Hoiberg, Dale and Clancy, Dan and Norvig, Peter and Orwant, Jon and Pinker, Steven and Nowak, Martin A. and Lieberman Aiden, Erez},
  journal={Science},
  volume={331},
  number={6014},
  pages={176--182},
  year={2011},
  publisher={American Association for the Advancement of Science}
}

@inproceedings{basile-etal-2020-diachronic,
    title = "A Diachronic {I}talian Corpus based on ``{L}{'}{U}nit{\`a}''",
    author = "Basile, Pierpaolo  and
      Caputo, Annalina  and
      Caselli, Tommaso  and
      Cassotti, Pierluigi  and
      Varvara, Rossella",
    editor = "Monti, Johanna  and
      Dell'Orletta, Felice  and
      Tamburini, Fabio",
    booktitle = "Proceedings of the Seventh Italian Conference on Computational Linguistics (CLiC-it 2020)",
    month = mar,
    year = "2020",
    address = "Bologna, Italy",
    publisher = "CEUR Workshop Proceedings",
    url = "https://aclanthology.org/2020.clicit-1.56/",
    pages = "369--374",
    ISBN = "979-12-80136-28-2"
}

@inproceedings{del2016tracing,
  title={Tracing metaphors in time through self-distance in vector spaces},
  author={Del Tredici, Marco and Nissim, Malvina and Zaninello, Andrea},
  booktitle={3rd Italian Conference on Computational Linguistics, CLiC-it 2016 and 5th Evaluation Campaign of Natural Language Processing and Speech Tools for Italian, EVALITA 2016},
  year={2016},
  url={https://arxiv.org/abs/1611.03279}
}

@inproceedings{10.1145/3178876.3185999,
author = {Rudolph, Maja and Blei, David},
title = {Dynamic Embeddings for Language Evolution},
year = {2018},
isbn = {9781450356398},
publisher = {International World Wide Web Conferences Steering Committee},
address = {Republic and Canton of Geneva, CHE},
url = {https://doi.org/10.1145/3178876.3185999},
doi = {10.1145/3178876.3185999},
booktitle = {Proceedings of the 2018 World Wide Web Conference},
pages = {1003–1011},
numpages = {9},
location = {Lyon, France},
series = {WWW '18}
}

@inproceedings{miletic-schulte-im-walde-2025-modeling,
    title = "Modeling the Evolution of {E}nglish Noun Compounds with Feature-Rich Diachronic Compositionality Prediction",
    author = "Mileti{\'c}, Filip  and
      Schulte im Walde, Sabine",
    editor = "Che, Wanxiang  and
      Nabende, Joyce  and
      Shutova, Ekaterina  and
      Pilehvar, Mohammad Taher",
    booktitle = "Proceedings of the 63rd Annual Meeting of the Association for Computational Linguistics (Volume 1: Long Papers)",
    month = jul,
    year = "2025",
    address = "Vienna, Austria",
    publisher = "Association for Computational Linguistics",
    url = "https://aclanthology.org/2025.acl-long.984/",
    doi = "10.18653/v1/2025.acl-long.984",
    pages = "20071--20092",
    ISBN = "979-8-89176-251-0",
}

@inproceedings{10.1145/2488388.2488416,
author = {Danescu-Niculescu-Mizil, Cristian and West, Robert and Jurafsky, Dan and Leskovec, Jure and Potts, Christopher},
title = {No country for old members: user lifecycle and linguistic change in online communities},
year = {2013},
isbn = {9781450320351},
publisher = {Association for Computing Machinery},
address = {New York, NY, USA},
url = {https://doi.org/10.1145/2488388.2488416},
doi = {10.1145/2488388.2488416},
booktitle = {Proceedings of the 22nd International Conference on World Wide Web},
pages = {307–318},
numpages = {12},
location = {Rio de Janeiro, Brazil},
series = {WWW '13}
}

@inproceedings{10.1145/3696410.3714716,
author = {Ananthasubramaniam, Aparna and Zhu, Yufei `Louise' and Jurgens, David and Romero, Daniel M.},
title = {Roles of Network and Identity in Hashtag Diffusion},
year = {2025},
isbn = {9798400712746},
publisher = {Association for Computing Machinery},
address = {New York, NY, USA},
url = {https://doi.org/10.1145/3696410.3714716},
doi = {10.1145/3696410.3714716},
booktitle = {Proceedings of the ACM on Web Conference 2025},
pages = {3233–3246},
numpages = {14},
location = {Sydney NSW, Australia},
series = {WWW '25}
}

@article{10.1371/journal.pone.0113114,
    doi = {10.1371/journal.pone.0113114},
    author = {Eisenstein, Jacob AND O'Connor, Brendan AND Smith, Noah A. AND Xing, Eric P.},
    journal = {PLOS ONE},
    publisher = {Public Library of Science},
    title = {Diffusion of Lexical Change in Social Media},
    year = {2014},
    month = {11},
    volume = {9},
    url = {https://doi.org/10.1371/journal.pone.0113114},
    pages = {1-13},
    number = {11},
}

@book{aitchison2001language,
  title={Language change: Progress or decay?},
  author={Aitchison, Jean},
  year={2001},
  publisher={Cambridge University Press}
}

@book{milroy2012,
    author = {Milroy, James AND Milroy, Leslie},
    title = {Authority in Language: Investigating Standard {E}nglish},
    publisher = "Routledge",
    year = 2012
}

@book{rogers1971communication,
  title={Communication of innovations: a cross-cultural approach},
  author={Rogers, Everett M. and Schoemaker, F. Floyd},
  publisher={The Free Press},
  year = 1971
}

@book{montgomery2008introduction,
  title={An introduction to language and society},
  author={Montgomery, Martin},
  year={2008},
  publisher={Routledge}
}

@incollection{lefkowitz2017anti,
    author = {Lefkowitz, Natalie and Hedgcock, John S.},
    title = {Anti-language: Linguistic innovation, identity construction, and group affiliation among emerging speech communities},
    year = {2017},
    booktitle={Multiple Perspectives on Language Play},  pages={347--376},
    publisher = {De Gruyter Mouton},
    editor={Nancy Bell}
}

@book{mcculloch2020because,
  title={Because Internet: Understanding the new rules of language},
  author={McCulloch, Gretchen},
  year={2020},
  publisher={Penguin}
}

@ARTICLE{10.3389/frai.2021.648583,
AUTHOR={Würschinger, Quirin},
TITLE={Social Networks of Lexical Innovation. Investigating the Social Dynamics of Diffusion of Neologisms on {T}witter},
JOURNAL={Frontiers in Artificial Intelligence},
VOLUME={4},
YEAR={2021},
URL={https://www.frontiersin.org/journals/artificial-intelligence/articles/10.3389/frai.2021.648583},
DOI={10.3389/frai.2021.648583},
}

@book{bird2009natural,
  title={Natural language processing with {P}ython},
  author={Bird, Steven and Klein, Ewan and Loper, Edward},
  year={2009},
  publisher={O'Reilly Media Inc.}
}

@article{de2021marcador,
  title={El marcador conversacional ``ahre'' en el habla de estudiantes adolescentes de {B}uenos {A}ires: un estudio de sus usos en dos g{\'e}neros conversacionales},
  author={De Luca, Natalia},
  journal={Revista Latinoamericana de Estudios del Discurso},
  volume={21},
  number={2},
  pages={27--48},
  year={2021},
  publisher={Asociaci{\'o}n Latinoamericana de Estudios del Discurso (ALED)},
  url={https://dialnet.unirioja.es/servlet/articulo?codigo=9127762}
}

@incollection{peppler2013social,
  title={Social media and creativity},
  author={Peppler, Kylie},
  booktitle={The Routledge International Handbook of Children, Adolescents and Media},
  pages={219--226},
  year={2013},
  publisher={Routledge}
}

@article{grieve2018mapping,
  title={Mapping lexical innovation on {A}merican social media},
  author={Grieve, Jack and Nini, Andrea and Guo, Diansheng},
  journal={Journal of English Linguistics},
  volume={46},
  number={4},
  pages={293--319},
  year={2018},
  publisher={SAGE Publications Sage CA: Los Angeles, CA}
}

@inproceedings{mccauley2006technical,
  title={Technical Combining Forms in the Third Edition of the {OED}: Word Formation in a Historical Dictionary},
  author={McCauley, Jane},
  booktitle={Selected proceedings of the 2005 Symposium on New Approaches in English Historical Lexis (HEL-LEX). Somerville, MA: Cascadilla Proceedings Project},
  pages={95--104},
  year={2006},
  url={https://www.lingref.com/cpp/hel-lex/2005/paper1350.pdf}
}

@inproceedings{bommasani-etal-2020-interpreting-2,
    title = "Interpreting Pretrained Contextualized Representations via Reductions to Static Embeddings",
    author = "Bommasani, Rishi  and
      Davis, Kelly  and
      Cardie, Claire",
    booktitle = "Proceedings of the 58th Annual Meeting of the Association for Computational Linguistics",
    month = jul,
    year = "2020",
    address = "Online",
    publisher = "Association for Computational Linguistics",
    url = "https://aclanthology.org/2020.acl-main.431/",
    pages = "4758--4781",
}

@inproceedings{bowern-2019-semantic-2,
    title = "Semantic Change and Semantic Stability: Variation is Key",
    author = "Bowern, Claire",
    editor = "Tahmasebi, Nina  and
      Borin, Lars  and
      Jatowt, Adam  and
      Xu, Yang",
    booktitle = "Proceedings of the 1st International Workshop on Computational Approaches to Historical Language Change",
    month = aug,
    year = "2019",
    address = "Florence, Italy",
    publisher = "Association for Computational Linguistics",
    url = "https://aclanthology.org/W19-4706/",
    pages = "48--55",
}

@inproceedings{haider-eger-2019-semantic-2,
    title = "Semantic Change and Emerging Tropes In a Large Corpus of {N}ew {H}igh {G}erman Poetry",
    author = "Haider, Thomas  and
      Eger, Steffen",
    editor = "Tahmasebi, Nina  and
      Borin, Lars  and
      Jatowt, Adam  and
      Xu, Yang",
    booktitle = "Proceedings of the 1st International Workshop on Computational Approaches to Historical Language Change",
    month = aug,
    year = "2019",
    address = "Florence, Italy",
    publisher = "Association for Computational Linguistics",
    url = "https://aclanthology.org/W19-4727/",
    pages = "216--222",
}

@inproceedings{hofmann-etal-2020-predicting-2,
    title = "Predicting the Growth of Morphological Families from Social and Linguistic Factors",
    author = {Hofmann, Valentin  and
      Pierrehumbert, Janet  and
      Sch{\"u}tze, Hinrich},
    editor = "Jurafsky, Dan  and
      Chai, Joyce  and
      Schluter, Natalie  and
      Tetreault, Joel",
    booktitle = "Proceedings of the 58th Annual Meeting of the Association for Computational Linguistics",
    month = jul,
    year = "2020",
    address = "Online",
    publisher = "Association for Computational Linguistics",
    url = "https://aclanthology.org/2020.acl-main.649/",
    pages = "7273--7283",
}

@inproceedings{ryskina-etal-2020-phonetic-2,
    title = "Phonetic and Visual Priors for Decipherment of Informal Romanization",
    author = "Ryskina, Maria  and
      Gormley, Matthew R.  and
      Berg-Kirkpatrick, Taylor",
    editor = "Jurafsky, Dan  and
      Chai, Joyce  and
      Schluter, Natalie  and
      Tetreault, Joel",
    booktitle = "Proceedings of the 58th Annual Meeting of the Association for Computational Linguistics",
    month = jul,
    year = "2020",
    address = "Online",
    publisher = "Association for Computational Linguistics",
    url = "https://aclanthology.org/2020.acl-main.737/",
    pages = "8308--8319",
}

@inproceedings{weissweiler-etal-2023-counting-2,
    title = "Counting the Bugs in {C}hat{GPT}'s Wugs: A Multilingual Investigation into the Morphological Capabilities of a Large Language Model",
    author = "Weissweiler, Leonie  and
      Hofmann, Valentin  and
      Kantharuban, Anjali  and
      Cai, Anna  and
      Dutt, Ritam  and
      Hengle, Amey  and
      Kabra, Anubha  and
      Kulkarni, Atharva  and
      Vijayakumar, Abhishek  and
      Yu, Haofei  and
      Schuetze, Hinrich  and
      Oflazer, Kemal  and
      Mortensen, David",
    editor = "Bouamor, Houda  and
      Pino, Juan  and
      Bali, Kalika",
    booktitle = "Proceedings of the 2023 Conference on Empirical Methods in Natural Language Processing",
    month = dec,
    year = "2023",
    address = "Singapore",
    publisher = "Association for Computational Linguistics",
    url = "https://aclanthology.org/2023.emnlp-main.401/",
    doi = "10.18653/v1/2023.emnlp-main.401",
    pages = "6508--6524",
}

@inproceedings{kulkarni2015statistically,
  title={Statistically significant detection of linguistic change},
  author={Kulkarni, Vivek and Al-Rfou, Rami and Perozzi, Bryan and Skiena, Steven},
  booktitle={Proceedings of the 24th International Conference on World Wide Web},
  pages={625--635},
  year={2015},
  doi={10.1145/2736277.2741627}
}

@ARTICLE{7511732,
  author={Zhang, Yating and Jatowt, Adam and Bhowmick, Sourav S. and Tanaka, Katsumi},
  journal={IEEE Transactions on Knowledge and Data Engineering}, 
  title={The Past is Not a Foreign Country: Detecting Semantically Similar Terms across Time}, 
  year={2016},
  volume={28},
  number={10},
  pages={2793-2807},
  doi={10.1109/TKDE.2016.2591008}
}

\appendix

\section{Differences from \citeposs{ryskina-etal-2020-new} experimental setup}
\label{sec:appendix-diff}

Our opertionalization of the hypotheses (\Sref{sec:setup}) builds on that of \citet{ryskina-etal-2020-new}, with several differences:

\begin{itemize}
    \item Since the differences in neighbourhood density between neologisms and controls can be dramatic, we measure it as $\log (|N_{\tau}(w)| + 1)$ (Eq. \ref{eq:density}) rather than the raw number of neighbours $|N_{\tau}(w)|$.
    \item We propose a more robust estimation for the frequency growth monotonicity measure: instead of estimating each neighbour word's growth monotonicity and averaging, we measure the growth of the neighbourhood as a whole (Eq. \ref{eq:growth}) and estimate the monotonicity of that (Eq. \ref{eq:growth-monotonicity}).
    \item To test the demand hypothesis more thoroughly, we also compute the regression slope of the frequency growth (Eq. \ref{eq:growth-slope}) as an additional metric.
\end{itemize}

We also use different criteria for neologism selection (\Sref{sec:selection}) and neologism--control pairing (\Sref{sec:control}), use different hyperparameters for Word2Vec (\Sref{sec:embeddings-word2vec}), and additionally repeat all experiments with contextual embeddings (\Sref{sec:embeddings-roberta}).

\section{Data preprocessing and filtering}

\subsection{Tweet preprocessing and tokenization}
\label{sec:appendix-tokenization}

At the preprocessing step, all tweets are lowercased and tokenized using NLTK's \texttt{TweetTokenizer} \cite{bird2009natural}. Using regular expression-based heuristics, we remove the following from all tokenized tweets: URLs; phone numbers; hashtags; tokens consisting exclusively of emoji, numbers, punctuation, or special characters; single-character tokens; and tokens longer than 50 characters. This leads to an extra 0.3\% of all tweets being filtered out entirely. We use this tokenized data to:
\begin{itemize}
    \item extract the vocabulary and estimate token frequencies, which are then used to select candidate neologisms (\Sref{sec:selection-twitter}) and to pair neologisms with controls (\Sref{sec:control});
    \item train the static Word2Vec models (\Sref{sec:embeddings-word2vec}).
\end{itemize}

Since RoBERTa uses its own tokenizer, we extract the contextual embeddings (\Sref{sec:embeddings-roberta}) using the original, non-preprocessed tweet text. However, some vocabulary words extracted by \texttt{TweetTokenizer} do not match RoBERTa's token boundaries. For example, in
\begin{quote}
    \texttt{RT@<USERNAME>:Parenting tip ...}
\end{quote}
\texttt{TweetTokenizer} considers \texttt{:P} to be a separate emoticon token, leading to \texttt{arenting} appearing in the historical vocabulary. Such tokenization mismatches are excluded from the historical vocabulary in all RoBERTa-based analyses.

\subsection{Automatic Twitter neologism filtering}
\label{sec:appendix-pos}

The initial list of neologism candidates obtained by thresholding the \Modern{} Twitter vocabulary by the first year of popular use (\Sref{sec:selection-twitter}) consists of 3554 words. First, we refine the resulting list using part-of-speech tagging: for each word, we sample 100 tweets containing it at random and run them through the Flair English POS tagger\footnote{\url{https://huggingface.co/flair/pos-english}}~\cite{akbik-etal-2018-contextual}. Any potential neologisms for which the most frequent tag was NNP or NNPS (proper nouns), FW (foreign), CD (number) or NFP (superfluous punctuation) are discarded. Second, we filter out rare variants by removing any words that occur fewer than 500 times in the entire corpus. Finally, we remove any words whose frequency distribution is sparse and sharply peaked (usually associated with auto-generated, templatic tweets). For example, \texttt{theweatherchannel} has 130K occurrences in 2016 and none in 2017--2021, and all 2016 occurrences follow the same template: 
\begin{quote}
    \texttt{Get Weather Updates from \underline{theweatherchannel} <TIMESTAMP>}
\end{quote}

The 938 words remaining after this step are additionally filtered manually, as described in the next section. The results with the non-filtered list of words are reported in \Sref{sec:appendix-full}.

\subsection{Manual neologism filtering}
\label{sec:appendix-filtering}

\paragraph{Published writing}

We look up each of the 1000 potential neologisms extracted by \citet{ryskina-etal-2020-new} in the Oxford English Dictionary (OED).\footnote{\url{https://www.oed.com/}} An OED entry typically provides etymological information as well as the usage statistics for each recorded distinct sense of the word. For each word, we note the year it was first used in its latest sense in any part of speech (e.g., \emph{icon} in the computing sense) or as a part of its newest OED-recorded compound (such as \emph{in-vitro fertilization} for \emph{in-vitro}). We discard any words for which this year is earlier than 1900. For words not found in OED entries (e.g., \emph{p-value}), we rely on either the first use date per the Merriam-Webster
dictionary\footnote{\url{https://www.merriam-webster.com/}} or the year of the word's first appearance in the OED quotation bank. Finally, we discard the nine remaining words not found in either dictionary.

\paragraph{Twitter}

Starting with the list of 938 neologism candidates, obtained as described in \Sref{sec:selection-twitter}, we look up each word's context in our Twitter dataset, as well in the variety of online sources.

We design the following filtering rules:

\begin{itemize}
    \item We discard foreign words for which no evidence of natural code-switched use was found in our corpus. These occur in tweets created by sharing from other websites (e.g., \emph{``... via YouTube''}):
    \begin{quote}
        \texttt{Need for Speed - Most Wanted Soundtrack (Full) <URL> \underline{przez} @YouTube}
    \end{quote}
    We keep foreign words used in code-switched contexts:
    \begin{quote}
        \texttt{Nha \underline{Bafethu} the Messi saga is difficult to accept}
    \end{quote}
    \item We discard proper names, including names of people, works of art, products, and companies:
    \begin{quote}
        \texttt{VOTE \underline{SUNGHOON} ON TIKTOK!}
    \end{quote}
    We keep derivations or abbreviations of product names that have gained common use (e.g., per Urban Dictionary):\footnote{\url{https://www.urbandictionary.com/}}
    \begin{quote}
        \texttt{I just found a bunch of old selfies/\underline{insta} posts}
    \end{quote}
    We also keep product names converted into verbs:
    \begin{quote}
        \texttt{using snapchat for the memories >{}>{}> actually \underline{snapchatting}}
    \end{quote}
    \item We discard typos:
    \begin{quote}
        \texttt{DM US ON OUR \underline{INSTRAGRAM} ACCOUNT}
    \end{quote}
    We use the Urban Dictionary to differentiate them from deliberate creative spellings, which we keep: 
    \begin{quote}
        \texttt{i \underline{literallt}\footnote{\url{https://www.urbandictionary.com/define.php?term=Literallt}} cant}
    \end{quote}
    \item We discard tokenization errors, such as \texttt{lled} being identified as a neologism candidate because of incorrectly splitting a token:
    \begin{quote}
        \texttt{i forgot to censor k!\underline{lled}}
    \end{quote}
    \item We discard words which have been in use prior to 2000 per the Oxford English Dictionary (OED) and have not gained new senses since (e.g., \texttt{unvaccinated}).
    We rely on both the OED and the Urban Dictionary for identifying senses: for example, the OED lists only the currency-related sense of \texttt{demonetization} (in use since 1795), while the Urban Dictionary lists its newer content-related sense.\footnote{\url{https://www.urbandictionary.com/define.php?term=Demonetization}}
\end{itemize}

\section{Neologism formation categorization}
\label{sec:appendix-word-formation}

We primarily rely on the Oxford English Dictionary (OED) to categorize neologisms by how they were created. For non-dictionary words or ones where the etymological information is not provided, we reconstruct it using  Wiktionary\footnote{\url{https://www.wiktionary.org/}} and other online resources. 

The statistics for both corpora are reported in \Tref{tab:formation}. We use the following categories:
\begin{itemize}
    \item Abbreviation: clippings or shortenings \emph{(information $\rightarrow$ info)}, acronyms \emph{(canola)}, and initialisms \emph{(aka)}. Following the OED labels, we also include clipped compounds \emph{(sitcom)} and loan-phrases shortened to single words \emph{(caffè latte $\rightarrow$ latte)}.
    \item Derivation: morphological derivatives, primarily words created by affixation \emph{(light $\rightarrow$ ultralight)}. Words that the OED annotates as created by back formation \emph{(grungy $\rightarrow$ grunge)} are included in this category.
    \item Borrowing: loanwords from other languages, including cases where a borrowed root is combined with an English affix (Greek \emph{`leptos' + ‑in $\rightarrow$ leptin}). For Twitter neologisms, we include foreign words used in code-switched contexts in this category.
    \item Compound: words combined in their entirety \emph{(bodysuit, business-to-business)} or words combined with affixes that the OED classifies as combining forms (such as \emph{bio-, tele-}; \citealp{mccauley2006technical}).
    \item Blend: words combined in a way that alters at least one of their original forms (e.g., \emph{cell(ular) + phone $\rightarrow$ cellphone, t(rans)- + am(ine) + oxy- + phen(ol) $\rightarrow$ tamoxifen}). Borrowed roots combined in such a way with English elements (Greek \emph{`ergon' + (econ)omics) $\rightarrow$ ergonomics}. The OED does not differentiate between blends and compounds, so we classify them manually. 
    \item Spelling: orthographic alterations of an existing word or phrase \emph{(gangster $\rightarrow$ gangsta)}. Orthographic renderings of pronunciation or accent \emph{(both of them $\rightarrow$ boffum, heart $\rightarrow$ hearteu)}, onomatopoeia \emph{(skrrt)}, and keysmash \cite[\emph{sksksk};][]{mcculloch2020because} are also included in this category.
    \item Sense: a commonly used word acquiring an new sense. While the OED provides both the original etymology (e.g., borrowing for \emph{analog}) and the timelines for the various senses (e.g., 1941– for \emph{analog} as `non-digital'), for many words (e.g., \emph{browser}) it is hard to say whether the new sense was transferred onto an existing form or the form itself was re-coined from scratch. We chose to use this label (in addition to the original OED-annotated category) primarily for neologisms acquiring figurative senses.
    \item POS conversion: an existing word taking on a new part of speech (e.g., \emph{aerobic}, adj. $\rightarrow$ \emph{aerobics}, noun). If a word has multiple OED entries for different parts of speech listing different etymologies (e.g., derivation for \emph{interface}, n. and conversion for \emph{interface}, v.), we use labels from all the entries where the first use of the most recent sense is in 1900 or later. Where the OED combines parts of speech in the same entry, we follow the dictionary and do not use this label.
\end{itemize}

Many Twitter neologisms do not appear in the OED, so we annotated them manually, aiming for consistency with the OED categorization. The same word can fall into multiple categories (e.g., \emph{bodycam} is both a compound and an abbreviation).

\section{Results for non-filtered neologisms}
\label{sec:appendix-full}

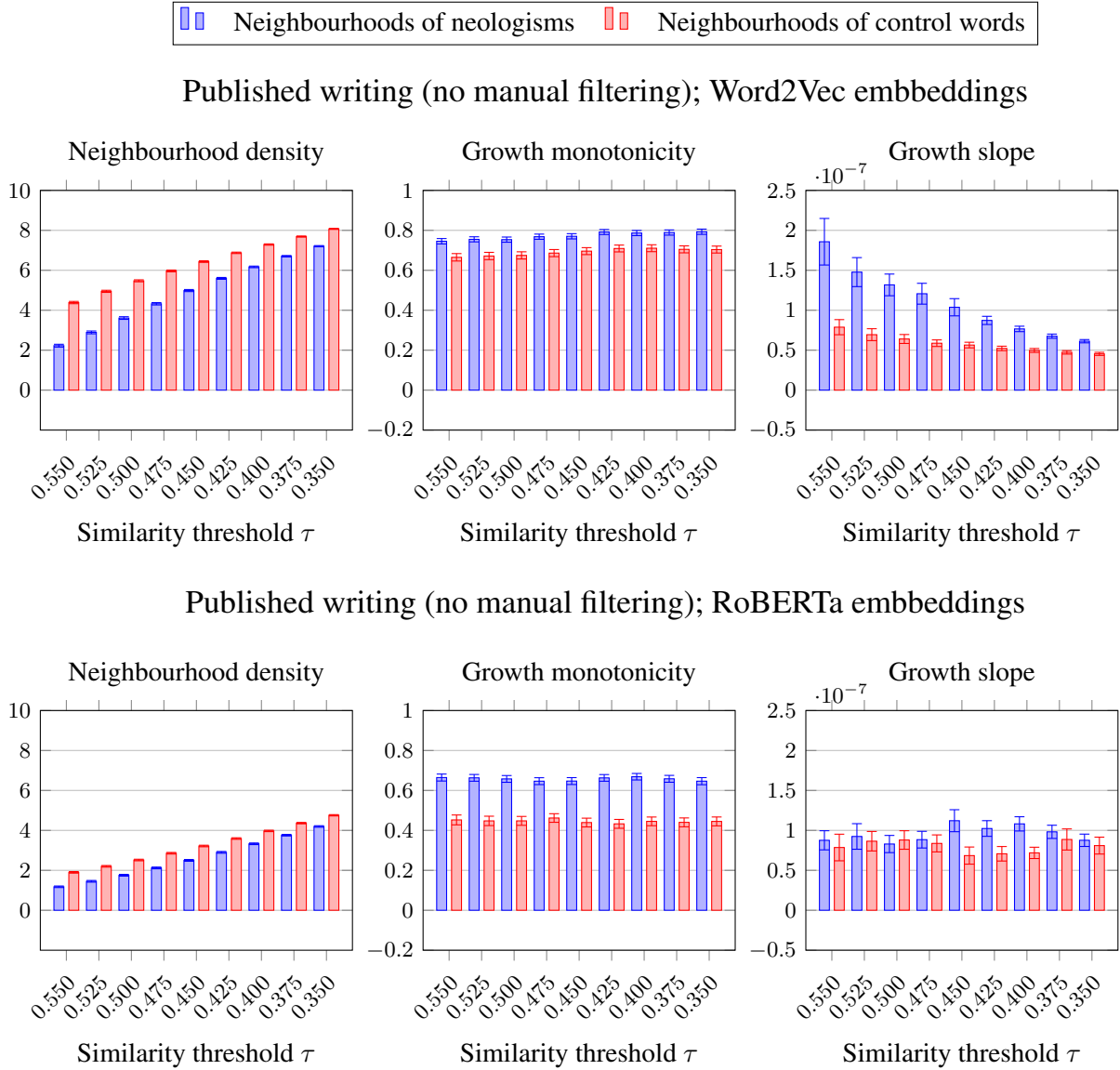
\begin{figure*}[t]
\centering
\begin{tikzpicture}
\pgfplotsset{small, every tick label/.append style={font=\footnotesize}}
\begin{groupplot}[group style = {group size = 3 by 2, horizontal sep = 1cm, vertical sep = 4cm}, width = 6cm, height = 5cm, ]
    \nextgroupplot[title = {Neighbourhood density},
        legend style = { column sep = 10pt, legend columns = -1, legend to name = grouplegend3},
        x tick label style={rotate=45,anchor=north east,/pgf/number format/.cd, fixed, fixed zerofill,
        precision=3},
        xlabel=\normalsize{Similarity threshold $\tau$},
        ybar,
        bar width = 4pt,
        ytick = {0,2,4,6,8,10},
        ymajorgrids=true,
        ymin=-2, ymax=10,
        x dir=reverse,
        xtick={0.55,0.525,0.5,0.475,0.45,0.425,0.4,0.375,0.35},
        ]
        \addplot+[error bars/.cd,y dir=both,y explicit]
        coordinates {
(0.55, 2.227370875705812) +- (0.0, 0.0700167885116358)
(0.525, 2.8911689834610104) +- (0.0, 0.0702587980047366)
(0.5, 3.609561104826406) +- (0.0, 0.066282457745009)
(0.475, 4.324092397307944) +- (0.0, 0.05951066592221077)
(0.45, 4.985548394273997) +- (0.0, 0.05259832745856169)
(0.425, 5.5958569362186665) +- (0.0, 0.04635200837647903)
(0.4, 6.170764250534684) +- (0.0, 0.040024475797009004)
(0.375, 6.705183737517399) +- (0.0, 0.03484109700831281)
(0.35, 7.206711192592487) +- (0.0, 0.030391807419115516)
    }; 
    \addlegendentry{Neighbourhoods of neologisms}%
        \addplot+[error bars/.cd,y dir=both,y explicit] 
        coordinates {
(0.55, 4.384474509808512) +- (0.0, 0.057814017691225426)
(0.525, 4.94766447347057) +- (0.0, 0.053537180369931034)
(0.5, 5.47419048146375) +- (0.0, 0.04937340231981819)
(0.475, 5.968758177406862) +- (0.0, 0.04453898754789022)
(0.45, 6.4363046395489105) +- (0.0, 0.04000258102638779)
(0.425, 6.873934645786085) +- (0.0, 0.036045394768548226)
(0.4, 7.290586950533) +- (0.0, 0.03222786806932718)
(0.375, 7.688882936405199) +- (0.0, 0.02870673921117721)
(0.35, 8.072142589143636) +- (0.0, 0.0253486609509093)
        }; \addlegendentry{Neighbourhoods of control words}
    \nextgroupplot[title={Growth monotonicity},
        x tick label style={rotate=45,anchor=north east,/pgf/number format/.cd, fixed, fixed zerofill,
        precision=3},        
        xlabel=\normalsize{Similarity threshold $\tau$},
        ybar,
        bar width = 4pt,
        ytick = {-0.2,0,0.2,0.4,0.6,0.8,1},
        ymajorgrids=true,
        ymin=-0.2, ymax=1,
        x dir=reverse,
        xtick={0.55,0.525,0.5,0.475,0.45,0.425,0.4,0.375,0.35},
        ]
        \addplot+[error bars/.cd,y dir=both,y explicit]
        coordinates {
(0.55, 0.7454948944621583) +- (0.0, 0.013399697279787879)
(0.525, 0.7550275113865913) +- (0.0, 0.012739803288790817)
(0.5, 0.7535212527319474) +- (0.0, 0.012968352089806645)
(0.475, 0.7681924676440028) +- (0.0, 0.013015950218371016)
(0.45, 0.7702767312296948) +- (0.0, 0.013164853646427522)
(0.425, 0.7917035669119398) +- (0.0, 0.012523985981881711)
(0.4, 0.7865923085860346) +- (0.0, 0.012874526180857273)
(0.375, 0.7891729713844409) +- (0.0, 0.013089831245765198)
(0.35, 0.7931533157006266) +- (0.0, 0.013206203954539085)
        };
        \addplot+[error bars/.cd,y dir=both,y explicit]
        coordinates {
(0.55, 0.6650108173555381) +- (0.0, 0.01858594193312938)
(0.525, 0.6715830982150859) +- (0.0, 0.018278191909784067)
(0.5, 0.6746208674162438) +- (0.0, 0.01795102043759834)
(0.475, 0.6861273950759792) +- (0.0, 0.017809439851477933)
(0.45, 0.6959574628038737) +- (0.0, 0.017455203402156543)
(0.425, 0.7091888271516344) +- (0.0, 0.01711775001818221)
(0.4, 0.7105256323512326) +- (0.0, 0.017307328803321403)
(0.375, 0.7048886337386122) +- (0.0, 0.017552915985418485)
(0.35, 0.7038771451417793) +- (0.0, 0.017386887017694554)
        };
    \nextgroupplot[title={Growth slope},
        x tick label style={rotate=45,anchor=north east,/pgf/number format/.cd, fixed, fixed zerofill,
        precision=3},        
        xlabel=\normalsize{Similarity threshold $\tau$},
        ybar,
        bar width = 4pt,
        ytick = {-0.00000005,0,0.00000005,0.0000001,0.00000015,0.0000002,0.00000025},
        ymajorgrids=true,
        ymin=-0.00000005, ymax=0.00000025,
        x dir=reverse,
        xtick={0.55,0.525,0.5,0.475,0.45,0.425,0.4,0.375,0.35},
        ]
        \addplot+[error bars/.cd,y dir=both,y explicit]
        coordinates {
(0.55, 1.8574817445236864e-07) +- (0.0, 2.9148020897526547e-08)
(0.525, 1.4773673659908188e-07) +- (0.0, 1.8152372912555444e-08)
(0.5, 1.3173482466918817e-07) +- (0.0, 1.3570971682950451e-08)
(0.475, 1.206797825207819e-07) +- (0.0, 1.304857167737059e-08)
(0.45, 1.0374658789397429e-07) +- (0.0, 1.0777806142936638e-08)
(0.425, 8.727918098734089e-08) +- (0.0, 5.120847748334761e-09)
(0.4, 7.662825286324826e-08) +- (0.0, 3.3914402424215798e-09)
(0.375, 6.754973452596391e-08) +- (0.0, 2.616946879523676e-09)
(0.35, 6.124284257899882e-08) +- (0.0, 2.0696595806434955e-09)
        };
        \addplot+[error bars/.cd,y dir=both,y explicit]
        coordinates {
(0.55, 7.881042955008114e-08) +- (0.0, 9.454157537215546e-09)
(0.525, 6.9425745281099e-08) +- (0.0, 7.438516179501367e-09)
(0.5, 6.40234237080194e-08) +- (0.0, 5.59399110537885e-09)
(0.475, 5.8743316982964504e-08) +- (0.0, 4.126909445280072e-09)
(0.45, 5.64334610038108e-08) +- (0.0, 3.5207869953150166e-09)
(0.425, 5.2237481447881645e-08) +- (0.0, 2.779317975396104e-09)
(0.4, 4.9685208623422825e-08) +- (0.0, 2.4193127180922846e-09)
(0.375, 4.729297250627872e-08) +- (0.0, 2.1818958941645923e-09)
(0.35, 4.539212118451909e-08) +- (0.0, 1.930281671684893e-09)
        };

    \nextgroupplot[title = {Neighbourhood density},
        x tick label style={rotate=45,anchor=north east,/pgf/number format/.cd, fixed, fixed zerofill,
        precision=3},
        xlabel=\normalsize{Similarity threshold $\tau$},
        ybar,
        bar width = 4pt,
        ytick = {0,2,4,6,8,10},
        ymajorgrids=true,
        ymin=-2, ymax=10,
        x dir=reverse,
        xtick={0.55,0.525,0.5,0.475,0.45,0.425,0.4,0.375,0.35},
        ]
        \addplot+[error bars/.cd,y dir=both,y explicit]
        coordinates {
(0.55, 1.1763197134752887) +- (0.0, 0.03898949576790377)
(0.525, 1.450596732158806) +- (0.0, 0.04209161812227929)
(0.5, 1.7534109820305335) +- (0.0, 0.04389550601057256)
(0.475, 2.119273254048045) +- (0.0, 0.04417404794977698)
(0.45, 2.4974212203668666) +- (0.0, 0.04428175468348098)
(0.425, 2.9004274963347334) +- (0.0, 0.043349817081860635)
(0.4, 3.3284218394005705) +- (0.0, 0.04130479836457082)
(0.375, 3.7582892585705023) +- (0.0, 0.03952837299131416)
(0.35, 4.196226876667518) +- (0.0, 0.03733404485585322)
    }; 
        \addplot+[error bars/.cd,y dir=both,y explicit] 
        coordinates {
(0.55, 1.8987960592027286) +- (0.0, 0.03944594306960866)
(0.525, 2.201406769881539) +- (0.0, 0.03976230906954784)
(0.5, 2.5165973146110674) +- (0.0, 0.03981658123116958)
(0.475, 2.8554054386440133) +- (0.0, 0.039716099105498344)
(0.45, 3.2182195278132726) +- (0.0, 0.039151969348772644)
(0.425, 3.591962742712004) +- (0.0, 0.03793925147191836)
(0.4, 3.9729729338386557) +- (0.0, 0.03637374501678504)
(0.375, 4.362910566290774) +- (0.0, 0.03477055893040869)
(0.35, 4.756469966381363) +- (0.0, 0.03291173779725905)
        };
    \nextgroupplot[title={Growth monotonicity},
        x tick label style={rotate=45,anchor=north east,/pgf/number format/.cd, fixed, fixed zerofill,
        precision=3},        
        xlabel=\normalsize{Similarity threshold $\tau$},
        ybar,
        bar width = 4pt,
        ytick = {-0.2,0,0.2,0.4,0.6,0.8,1},
        ymajorgrids=true,
        ymin=-0.2, ymax=1,
        x dir=reverse,
        xtick={0.55,0.525,0.5,0.475,0.45,0.425,0.4,0.375,0.35},
        ]
        \addplot+[error bars/.cd,y dir=both,y explicit]
        coordinates {
(0.55, 0.6644822224628364) +- (0.0, 0.017429974038664364)
(0.525, 0.6630241894515349) +- (0.0, 0.016877810958531823)
(0.5, 0.6576348848440142) +- (0.0, 0.016935678337876972)
(0.475, 0.6463597479605816) +- (0.0, 0.017186058791306465)
(0.45, 0.6468464441625378) +- (0.0, 0.017217711192792483)
(0.425, 0.6627506706773244) +- (0.0, 0.016675707095548425)
(0.4, 0.6686276520187439) +- (0.0, 0.01630547822880271)
(0.375, 0.6581501973525784) +- (0.0, 0.017147297968555856)
(0.35, 0.646349810020718) +- (0.0, 0.017703580301201647)
        };
        \addplot+[error bars/.cd,y dir=both,y explicit]
        coordinates {
(0.55, 0.45233579323531503) +- (0.0, 0.02465826719933945)
(0.525, 0.44750759451467614) +- (0.0, 0.02346639592128169)
(0.5, 0.44758238529071265) +- (0.0, 0.022295671364499992)
(0.475, 0.4618955845856179) +- (0.0, 0.021553780106913063)
(0.45, 0.43916837669203895) +- (0.0, 0.02195997817845355)
(0.425, 0.43265841632220886) +- (0.0, 0.022201011125113378)
(0.4, 0.44512783777221004) +- (0.0, 0.021727327816384846)
(0.375, 0.4401451051637176) +- (0.0, 0.022153093359653476)
(0.35, 0.4448166397241822) +- (0.0, 0.02223054492609185)
        };
    \nextgroupplot[title={Growth slope},
        x tick label style={rotate=45,anchor=north east,/pgf/number format/.cd, fixed, fixed zerofill,
        precision=3},        
        xlabel=\normalsize{Similarity threshold $\tau$},
        ybar,
        bar width = 4pt,
        ytick = {-0.00000005,0,0.00000005,0.0000001,0.00000015,0.0000002,0.00000025},
        ymajorgrids=true,
        ymin=-0.00000005, ymax=0.00000025,
        x dir=reverse,
        xtick={0.55,0.525,0.5,0.475,0.45,0.425,0.4,0.375,0.35},
        ]
        \addplot+[error bars/.cd,y dir=both,y explicit]
        coordinates {
(0.55, 8.75591711043025e-08) +- (0.0, 1.1961453216409717e-08)
(0.525, 9.237344078959538e-08) +- (0.0, 1.601747123614165e-08)
(0.5, 8.290547163626963e-08) +- (0.0, 1.0751069605384991e-08)
(0.475, 8.837887317846794e-08) +- (0.0, 1.0429977472609073e-08)
(0.45, 1.1209586439422663e-07) +- (0.0, 1.3766287312698835e-08)
(0.425, 1.0232729748850802e-07) +- (0.0, 9.806660621538973e-09)
(0.4, 1.0813522313314979e-07) +- (0.0, 8.902424103213736e-09)
(0.375, 9.815961822723466e-08) +- (0.0, 8.259802571275314e-09)
(0.35, 8.759363004092186e-08) +- (0.0, 7.723876656194463e-09)
        };
        \addplot+[error bars/.cd,y dir=both,y explicit]
        coordinates {
(0.55, 7.856024687764746e-08) +- (0.0, 1.6652063584704828e-08)
(0.525, 8.643484678518697e-08) +- (0.0, 1.2201781731042906e-08)
(0.5, 8.793080919651277e-08) +- (0.0, 1.1594761662563116e-08)
(0.475, 8.368088750780505e-08) +- (0.0, 1.0644206635200562e-08)
(0.45, 6.843497717291333e-08) +- (0.0, 1.0740323323794038e-08)
(0.425, 7.06977964991686e-08) +- (0.0, 9.073301205048993e-09)
(0.4, 7.18212812774741e-08) +- (0.0, 7.087668770353285e-09)
(0.375, 8.860655021086033e-08) +- (0.0, 1.3195155279852056e-08)
(0.35, 8.094025016722111e-08) +- (0.0, 1.0570545909867608e-08)
        };
\end{groupplot}

\node at (8cm, 5.8cm) {\ref*{grouplegend3}}; 
\node (title) at (8cm, 4.8cm) {\larger{Published writing (no manual filtering); Word2Vec embbeddings}};
\node (title) at (8cm, -2.5cm) {\larger{Published writing (no manual filtering); RoBERTa embbeddings}};

\end{tikzpicture}
\caption{Experimental comparison between the neighbourhoods of neologisms (blue bars) and control words (red bars) in the \textbf{published writing} domain. Results are reported for 755 neologism--control pairs created from the original, \textbf{non-filtered neologism list} of 1000 candidate neologisms. The three plots in each row correspond to three measures: the number of \Historical{} neighbours a word has (left), how monotonically these neighbours grow in frequency (centre), and the linear regression slope of their growth (right). The x-axis on all plots corresponds to the neighbourhood size (defined by the cosine similarity threshold $\tau$). The top and bottom rows show the results with the static Word2Vec embeddings and the contextual RoBERTa embeddings respectively. Error bars represent standard error over words.}
\label{fig:coha_figure_full}
\end{figure*}
\begin{figure*}[t]
\centering
\begin{tikzpicture}
\pgfplotsset{small, every tick label/.append style={font=\footnotesize}}
\begin{groupplot}[group style = {group size = 3 by 2, horizontal sep = 1cm, vertical sep = 4cm}, width = 6cm, height = 5cm, ]
    \nextgroupplot[title = {Neighbourhood density},
        legend style = { column sep = 10pt, legend columns = -1, legend to name = grouplegend2},
        x tick label style={rotate=45,anchor=north east,/pgf/number format/.cd, fixed, fixed zerofill,
        precision=3},
        xlabel=\normalsize{Similarity threshold $\tau$},
        ybar,
        bar width = 4pt,
        ytick = {0,2,4,6,8,10},
        ymajorgrids=true,
        ymin=-2, ymax=10,
        x dir=reverse,
        xtick={0.55,0.525,0.5,0.475,0.45,0.425,0.4,0.375,0.35},
        ]
        \addplot+[error bars/.cd,y dir=both,y explicit]
        coordinates {
(0.55, 1.3002355548198146) +- (0.0, 0.06308807255051271)
(0.525, 1.806633995540846) +- (0.0, 0.06983720245644737)
(0.5, 2.4640959601293306) +- (0.0, 0.07155478488189118)
(0.475, 3.1414985265741797) +- (0.0, 0.07008714706204482)
(0.45, 3.8575070591287983) +- (0.0, 0.0640957740162589)
(0.425, 4.526749238996981) +- (0.0, 0.057703673197313295)
(0.4, 5.145546138099892) +- (0.0, 0.0517597267624777)
(0.375, 5.7294067248663465) +- (0.0, 0.04700671595149519)
(0.35, 6.300675646118412) +- (0.0, 0.04195042372016922)
    }; 
    \addlegendentry{Neighbourhoods of neologisms}%
        \addplot+[error bars/.cd,y dir=both,y explicit] 
        coordinates {
(0.55, 3.282879490127028) +- (0.0, 0.08890446292137326)
(0.525, 3.8344741696662186) +- (0.0, 0.08550534052411511)
(0.5, 4.377069382862057) +- (0.0, 0.0804314247704162)
(0.475, 4.899079762013443) +- (0.0, 0.07475406848871627)
(0.45, 5.410653571458046) +- (0.0, 0.06848902882676824)
(0.425, 5.902033361926056) +- (0.0, 0.06265499205718292)
(0.4, 6.379232049196317) +- (0.0, 0.05709487437628991)
(0.375, 6.85017231344453) +- (0.0, 0.051412359122716626)
(0.35, 7.3110816620818255) +- (0.0, 0.045638706009472826)
        }; \addlegendentry{Neighbourhoods of control words}
    \nextgroupplot[title={Growth monotonicity},
        x tick label style={rotate=45,anchor=north east,/pgf/number format/.cd, fixed, fixed zerofill,
        precision=3},        
        xlabel=\normalsize{Similarity threshold $\tau$},
        ybar,
        bar width = 4pt,
        ytick = {-0.2,0,0.2,0.4,0.6,0.8,1},
        ymajorgrids=true,
        ymin=-0.2, ymax=1,
        x dir=reverse,
        xtick={0.55,0.525,0.5,0.475,0.45,0.425,0.4,0.375,0.35},
        ]
        \addplot+[error bars/.cd,y dir=both,y explicit]
        coordinates {
(0.55, 0.5094846754410234) +- (0.0, 0.03615683772365837)
(0.525, 0.4526599342109991) +- (0.0, 0.033906809993821944)
(0.5, 0.4208126247878505) +- (0.0, 0.0321762445824789)
(0.475, 0.367255148093966) +- (0.0, 0.03135820168929623)
(0.45, 0.32755328354880997) +- (0.0, 0.03116364578918506)
(0.425, 0.3003685542410498) +- (0.0, 0.030867254804755283)
(0.4, 0.2737162968828194) +- (0.0, 0.03131595218238992)
(0.375, 0.26496090154346125) +- (0.0, 0.03146562549283303)
(0.35, 0.25099778270509976) +- (0.0, 0.031145262104427685)
        };
        \addplot+[error bars/.cd,y dir=both,y explicit]
        coordinates {
(0.55, 0.4412868771142612) +- (0.0, 0.0354795129997021)
(0.525, 0.39713393587090917) +- (0.0, 0.0335659944240301)
(0.5, 0.3668418611721658) +- (0.0, 0.031529304653950284)
(0.475, 0.33850766653650743) +- (0.0, 0.03131719678927901)
(0.45, 0.32033411977090004) +- (0.0, 0.031163034108763296)
(0.425, 0.2979955456570155) +- (0.0, 0.03148856567849604)
(0.4, 0.2611973392461197) +- (0.0, 0.031796885054534166)
(0.375, 0.24390243902439024) +- (0.0, 0.03253672904831439)
(0.35, 0.2164079822616408) +- (0.0, 0.03328737266986089)
        };
    \nextgroupplot[title={Growth slope},
        x tick label style={rotate=45,anchor=north east,/pgf/number format/.cd, fixed, fixed zerofill,
        precision=3},        
        xlabel=\normalsize{Similarity threshold $\tau$},
        ybar,
        bar width = 4pt,
        ytick = {-0.0000005,0,0.0000005,0.000001,0.0000015,0.000002,0.0000025},
        ymajorgrids=true,
        ymin=-0.0000005, ymax=0.0000025,
        x dir=reverse,
        xtick={0.55,0.525,0.5,0.475,0.45,0.425,0.4,0.375,0.35},
        ]
        \addplot+[error bars/.cd,y dir=both,y explicit]
        coordinates {
(0.55, 2.9650679300283753e-06) +- (0.0, 8.557665296110635e-07)
(0.525, 1.2481999802685038e-06) +- (0.0, 4.950694252078003e-07)
(0.5, 1.5028041223404784e-06) +- (0.0, 4.792954689294306e-07)
(0.475, 1.200832903259571e-06) +- (0.0, 4.689484160151399e-07)
(0.45, 8.594661504010553e-07) +- (0.0, 2.434758404687081e-07)
(0.425, 7.810884895569136e-07) +- (0.0, 1.934561439278209e-07)
(0.4, 5.494276214854014e-07) +- (0.0, 1.1095310478250516e-07)
(0.375, 3.895839698387298e-07) +- (0.0, 7.98209363543481e-08)
(0.35, 3.618005168747938e-07) +- (0.0, 6.167308738013862e-08)
        };
        \addplot+[error bars/.cd,y dir=both,y explicit]
        coordinates {
(0.55, 1.1067735301351495e-06) +- (0.0, 3.7510333619830444e-07)
(0.525, 7.160468846558983e-07) +- (0.0, 2.4331544847409264e-07)
(0.5, 3.5090763975600356e-07) +- (0.0, 1.7045643857209452e-07)
(0.475, 3.0436742602093886e-07) +- (0.0, 1.0737337916492644e-07)
(0.45, 3.364438075107511e-07) +- (0.0, 9.053838922561197e-08)
(0.425, 3.389313034027557e-07) +- (0.0, 8.293168571701236e-08)
(0.4, 3.554850636878152e-07) +- (0.0, 7.814359423974927e-08)
(0.375, 3.510857975055901e-07) +- (0.0, 6.552444619361686e-08)
(0.35, 3.4587755063546426e-07) +- (0.0, 6.943881395647627e-08)
        };

    \nextgroupplot[title = {Neighbourhood density},
        x tick label style={rotate=45,anchor=north east,/pgf/number format/.cd, fixed, fixed zerofill,
        precision=3},
        xlabel=\normalsize{Similarity threshold $\tau$},
        ybar,
        bar width = 4pt,
        ytick = {0,2,4,6,8,10},
        ymajorgrids=true,
        ymin=-2, ymax=10,
        x dir=reverse,
        xtick={0.55,0.525,0.5,0.475,0.45,0.425,0.4,0.375,0.35},
        ]
        \addplot+[error bars/.cd,y dir=both,y explicit]
        coordinates {
(0.55, 1.6679085343630895) +- (0.0, 0.06364451517463543)
(0.525, 1.9892209341526677) +- (0.0, 0.06562381569531894)
(0.5, 2.3555913452567534) +- (0.0, 0.06681612565726926)
(0.475, 2.7166358203960685) +- (0.0, 0.06770241030035822)
(0.45, 3.09736814408701) +- (0.0, 0.06780553932458619)
(0.425, 3.514895291374655) +- (0.0, 0.06505005487209885)
(0.4, 3.941262606519853) +- (0.0, 0.06165697746257346)
(0.375, 4.369177586046593) +- (0.0, 0.058497177819914425)
(0.35, 4.811859679436752) +- (0.0, 0.05492614285937289)
    }; 
        \addplot+[error bars/.cd,y dir=both,y explicit] 
        coordinates {
(0.55, 2.406416695094463) +- (0.0, 0.07203046941808464)
(0.525, 2.713676531503274) +- (0.0, 0.07119402051956875)
(0.5, 3.03182000918199) +- (0.0, 0.06974285143136484)
(0.475, 3.3538495276942415) +- (0.0, 0.0689712536421519)
(0.45, 3.698346148652725) +- (0.0, 0.06697529690984297)
(0.425, 4.054534579243375) +- (0.0, 0.06400431255268449)
(0.4, 4.4292059822765255) +- (0.0, 0.0602475650712007)
(0.375, 4.805106788946074) +- (0.0, 0.057259196133465445)
(0.35, 5.194755776802603) +- (0.0, 0.05359600397778129)
        };
    \nextgroupplot[title={Growth monotonicity},
        x tick label style={rotate=45,anchor=north east,/pgf/number format/.cd, fixed, fixed zerofill,
        precision=3},        
        xlabel=\normalsize{Similarity threshold $\tau$},
        ybar,
        bar width = 4pt,
        ytick = {-0.2,0,0.2,0.4,0.6,0.8,1},
        ymajorgrids=true,
        ymin=-0.2, ymax=1,
        x dir=reverse,
        xtick={0.55,0.525,0.5,0.475,0.45,0.425,0.4,0.375,0.35},
        ]
        \addplot+[error bars/.cd,y dir=both,y explicit]
        coordinates {
(0.55, 0.36423063156752283) +- (0.0, 0.03262642094092331)
(0.525, 0.36612989448177935) +- (0.0, 0.030168701588664776)
(0.5, 0.3428793443905104) +- (0.0, 0.02967723789547145)
(0.475, 0.3394492477340341) +- (0.0, 0.029424213001287662)
(0.45, 0.3300765426864812) +- (0.0, 0.029503759783937416)
(0.425, 0.35325751078656203) +- (0.0, 0.028957054771639068)
(0.4, 0.34633072387909225) +- (0.0, 0.02870137754033242)
(0.375, 0.3589502174843085) +- (0.0, 0.028107353486880475)
(0.35, 0.32682926829268294) +- (0.0, 0.028821422278087656)
        };
        \addplot+[error bars/.cd,y dir=both,y explicit]
        coordinates {
(0.55, 0.41900499757109866) +- (0.0, 0.03190141160986372)
(0.525, 0.40370164002233105) +- (0.0, 0.03122049348560503)
(0.5, 0.3440962886350849) +- (0.0, 0.03115723341059501)
(0.475, 0.31865666946104204) +- (0.0, 0.030097002407555654)
(0.45, 0.2940875731892633) +- (0.0, 0.029629014325178074)
(0.425, 0.29404732173957304) +- (0.0, 0.02926573426294979)
(0.4, 0.2848291026018806) +- (0.0, 0.028929721474671002)
(0.375, 0.24722254234168742) +- (0.0, 0.029840285833315792)
(0.35, 0.23414634146341462) +- (0.0, 0.030063916992133538)
        };
    \nextgroupplot[title={Growth slope},
        x tick label style={rotate=45,anchor=north east,/pgf/number format/.cd, fixed, fixed zerofill,
        precision=3},        
        xlabel=\normalsize{Similarity threshold $\tau$},
        ybar,
        bar width = 4pt,
        ytick = {-0.0000005,0,0.0000005,0.000001,0.0000015,0.000002,0.0000025},
        ymajorgrids=true,
        ymin=-0.0000005, ymax=0.0000025,
        x dir=reverse,
        xtick={0.55,0.525,0.5,0.475,0.45,0.425,0.4,0.375,0.35},
        ]
        \addplot+[error bars/.cd,y dir=both,y explicit]
        coordinates {
(0.55, 6.82051028916101e-08) +- (0.0, 1.0716729610252121e-07)
(0.525, 4.02644647594121e-08) +- (0.0, 8.024711820799571e-08)
(0.5, -4.625570160374662e-08) +- (0.0, 1.0179302180824854e-07)
(0.475, 1.4742768461094925e-08) +- (0.0, 8.408598257732143e-08)
(0.45, 7.405585628985148e-08) +- (0.0, 1.276522640912915e-07)
(0.425, 1.0754442444755636e-07) +- (0.0, 1.0654341936376309e-07)
(0.4, 9.103835062602266e-08) +- (0.0, 7.321741324042064e-08)
(0.375, 1.1461572586559783e-07) +- (0.0, 5.663627927766061e-08)
(0.35, 9.321198735588404e-08) +- (0.0, 4.500190007258028e-08)
        };
        \addplot+[error bars/.cd,y dir=both,y explicit]
        coordinates {
(0.55, 7.069380563017984e-07) +- (0.0, 4.976157230788475e-07)
(0.525, 5.305149638250605e-07) +- (0.0, 3.2110626712539344e-07)
(0.5, 4.2379178758763894e-07) +- (0.0, 2.518075683105195e-07)
(0.475, 3.1330526787244926e-07) +- (0.0, 2.075090005365859e-07)
(0.45, 1.511970896181857e-07) +- (0.0, 2.0820421544035047e-07)
(0.425, 4.5353406995680953e-07) +- (0.0, 3.7741586208803e-07)
(0.4, 3.8202334571361683e-07) +- (0.0, 2.99635368188739e-07)
(0.375, 3.4956160315545083e-07) +- (0.0, 2.14558028819857e-07)
(0.35, 2.31559449979576e-07) +- (0.0, 1.497170824183368e-07)
        };
\end{groupplot}

\node at (8cm, 5.8cm) {\ref*{grouplegend2}}; 
\node (title) at (8cm, 4.8cm) {\larger{Tweets (no manual filtering); Word2Vec embbeddings}};
\node (title) at (8cm, -2.5cm) {\larger{Tweets (no manual filtering); RoBERTa embbeddings}};

\end{tikzpicture}
\caption{Experimental comparison between the neighbourhoods of neologisms (blue bars) and control words (red bars) in the \textbf{Twitter} domain. Results are reported for 451 neologism--control pairs created from the original, \textbf{non-filtered neologism list} of 938 candidate neologisms. The three plots in each row correspond to three measures: the number of \Historical{} neighbours a word has (left), how monotonically these neighbours grow in frequency (centre), and the linear regression slope of their growth (right). The x-axis on all plots corresponds to the neighbourhood size (defined by the cosine similarity threshold $\tau$). The top and bottom rows show the results with the static Word2Vec embeddings and the contextual RoBERTa embeddings respectively. Error bars represent standard error over words.}
\label{fig:twitter_figure_full}
\end{figure*}
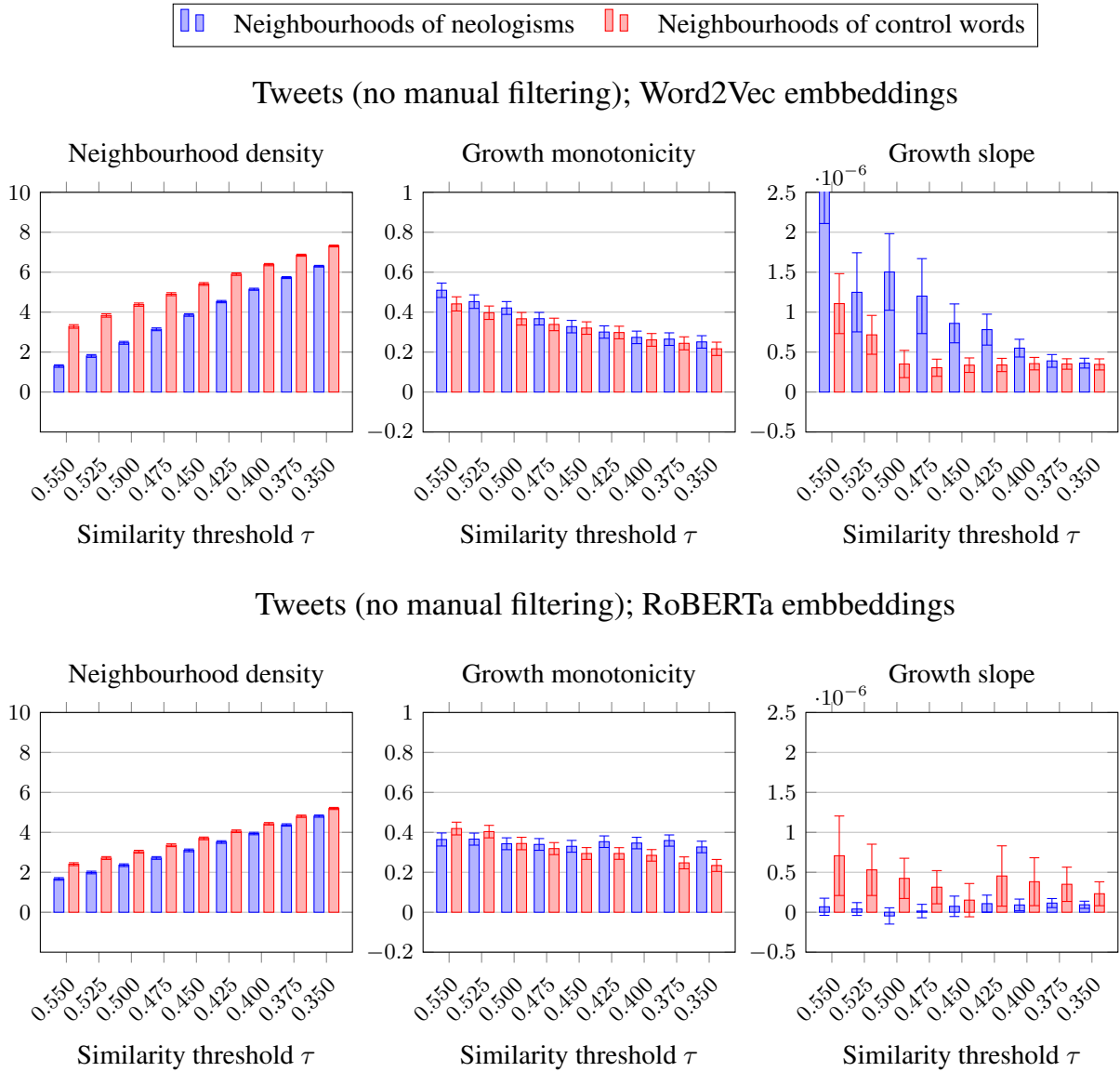

\Fref{fig:coha_figure_full} shows the results for the non-filtered list of published writing neologisms (755 neologism--control pairs). No suitable control words were found for the remaining 245 neologism candidates.

\Fref{fig:twitter_figure_full} shows the results for the non-filtered list of Twitter neologisms (451 neologism--control pairs). No suitable control words were found for the remaining 487 neologism candidates.

\end{document}